\documentclass{article} \usepackage{iclr2027_conference,times}
\usepackage{mathtools}
\usepackage{enumitem}
\usepackage{amsthm}
\usepackage{amssymb}

\usepackage{amsmath,amsfonts,bm}

\def\eqref#1{equation~\ref{#1}}

\def\1{\bm{1}}

\DeclareMathAlphabet{\mathsfit}{\encodingdefault}{\sfdefault}{m}{sl}
\SetMathAlphabet{\mathsfit}{bold}{\encodingdefault}{\sfdefault}{bx}{n}

\DeclareMathOperator{\Tr}{Tr}

\newtheorem{theorem}{Theorem}
\newtheorem{proposition}{Proposition}
\newtheorem{corollary}[theorem]{Corollary}
\newtheorem{lemma}[theorem]{Lemma}
\newtheorem{definition}{Definition}
\newtheorem{remark}{Remark}

\newcommand{\supp}{\operatorname{supp}}

\usepackage{hyperref}
\usepackage{url}

\renewcommand{\eqref}[1]{\textup{(\ref{#1})}}

\usepackage{nicefrac}

\title{ 
      Functional Compression as Occam's Razor:\\
      {\Large Algebraic Complexity from Sensitivity Minimization}
   }

\author{~~~~~~~~~~~~~~~
  Dongsung Huh \\ \\
          \And
  Lior Horesh \\ \\
        \And
  Halyun Jeong \\ \\
      }

\iclrfinalcopy 
\begin{document}

\maketitle
\vspace{-2.5em}

\begin{abstract}
Discovering discrete algebraic structures from data is a fundamental challenge in machine learning. We ask how continuous learning objectives can favor functional compression: representing data through shared algebraic composition rather than low-rank representations. We analyze the global optima of a canonical factorization model for algebraic completion (recovering finite binary operations) that demonstrated a preference for group operations. We measure algebraic complexity by the infimum of parameter sensitivity under exact reconstruction. We establish a universal complexity floor, whose attainment forces geometric alignment and unitarity of the latent representations. At the floor, input representations compose into the corresponding output representations; their reuse across the operation table propagates these local identities to enforce a globally shared algebraic structure. If the floor is not attained, a positive gap separates it from the feasible infimum. For cancellative operations, floor attainment forces the model's latent representation to be an injective homomorphism, equivalent to the target being isotopic to a group. In this case, every global minimizer is equivalent to the underlying group's unitary left-regular representation, up to isotopy and unitary gauge, resolving the central conjecture of Huh (2025). For non-cancellative operations, floor attainment forces exact structural redundancies (identical rows or columns) and is equivalent to a unitary factorization of the target's row- and column-reduced table with orthonormal output matrices, yielding the fewest distinct input matrices compatible with exact reconstruction. The resulting cancellative table then embeds in a group-isotope table. This analysis reveals a structural bias toward functional compression: at the sensitivity floor, algebraic composition of latent representations implements the target operation.
\end{abstract}

\section{Introduction}
\label{sec:intro}

Discovering algebraic rules from data---especially group structure---is a fundamental challenge in machine learning. This capability is critical for advancing algorithmic reasoning, mathematical problem-solving, and logic-based tasks \citep{powerGrokkingGeneralizationOverfitting2022, velickovic2021neural, lample2019deep}, which fundamentally rely on discovering and applying exact discrete rules. Furthermore, it holds profound implications for Geometric Deep Learning \citep{bronsteinGeometricDeepLearning2021}: automating the discovery of underlying symmetry groups \citep{benton2020learning, dehmamy2021automatic, finzi2021practical, chau2020disentangling, sanbornBispectralNeuralNetworks2023} can enable systematic generalization in domains where the true symmetry group is unknown. 

We consider \emph{algebraic completion} tasks---the recovery of an unknown binary operation on a finite set from partial observations \citep{powerGrokkingGeneralizationOverfitting2022}---as the canonical formalization for learning discrete algebraic structures.
Standard deep learning models often struggle in this regime, exhibiting significantly delayed generalization known as ``grokking'', especially with less training data \citep{powerGrokkingGeneralizationOverfitting2022, nanda2023progress, gromov2023grokking}.
Furthermore, while \citet{powerGrokkingGeneralizationOverfitting2022} reported that models learn and generalize more easily on ``simple'' operations (e.g., modular addition) than on ``complex'' ones (e.g., modular polynomials), how to measure operation-level complexity and connect it to a differentiable objective remains unresolved. 
{ Resolving this gap could guide the design of training objectives for more effective learning and generalization on symbolic tasks.}

Recently, \citet{huh2025discovering} demonstrated that a canonical factorization model with 
\textbf{ sensitivity minimization}---penalizing changes in model prediction under parameter perturbations---induces a systematic bias toward recovering group operations in algebraic completion, 
yielding improved sample efficiency and convergence speed over standard deep learning models. 
{ They defined the theoretical infimum of this sensitivity objective under exact target reconstruction as a measure of complexity, which we term \emph{algebraic complexity}. 
Furthermore, their empirical result (Figure~6) shows that lower algebraic complexity
is associated with better generalization from partial observations.}
However, the mechanism by which this objective favors group structure has remained unexplained.

In this work, we investigate the global minimum of the sensitivity objective to characterize algebraic complexity and explain its preference for group structure.

\subsection{{Background:} { Generalization in Matrix vs. Algebraic Completion}}
\label{sec:related_work}

\paragraph{Matrix Completion  via Rank Minimization.}
Matrix completion, the recovery of missing entries from partial observations, has long provided a fertile ground for developing rigorous theories of generalization. 
{ \emph{Rank} provides a natural notion of \emph{ geometric complexity} in matrices, where a lower rank indicates stronger linear dependencies among rows and columns. }
Matrix completion exploits these dependencies: 
classical works established recovery guarantees using explicit rank constraints or nuclear norm minimization \citep{candesExactMatrixCompletion2009, candesPowerConvexRelaxation2010}.
{ Moreover, a bias toward lower \emph{ geometric complexity} is readily induced by the differentiable framework of deep learning}, either explicitly through $\ell_2$ regularization \citep{shang2020unified,dai2021representation} or implicitly through small initialization \citep{saxeExactSolutionsNonlinear2014,huhLowRankSimplicityBias2023,jacot2023}. Deep matrix-factorization networks leverage this bias to achieve strong recovery in matrix completion \citep{gunasekarImplicitRegularizationMatrix2017,aroraImplicitRegularizationDeep2019a}.

\paragraph{Algebraic Completion.}
Algebraic completion provides a distinct ground for generalization theory for symbolic tasks. 
The entries of a binary operation table are discrete symbols, lacking any intrinsic meaning or numerical magnitudes. 
Because there is no inherent notion of proximity between these symbols, generalization must exploit the \emph{algebraic regularities} of the operation. 
{ However, which aspects of an operation's algebraic structure give rise to regularity, and how that regularity is reflected in its \emph{ algebraic complexity}, remain uncharacterized.}
In this work, we close this gap by addressing a fundamental question: what makes one operation algebraically simpler than another, and how can a corresponding simplicity bias be induced within a differentiable framework?

{
\emph{Functional compression} expresses an operation table through a shared composition rule rather than through rank reduction.
The factorization model represents symbols by matrices and can exactly reconstruct every finite binary operation table.
Our analysis characterizes when the sensitivity objective selects representations that realize this composition through matrix multiplication.
}

\subsection{Our Contributions}
\label{sec:contributions}

In this work, we propose the constrained infimum of the sensitivity objective $\mathcal{H}(\Theta)$ under exact reconstruction of a target operation table $\delta$ as a measure of \emph{algebraic complexity}, denoted as $\mathcal H_{\inf}(\delta)$. 
{ Our results connect sensitivity-floor attainment to
\emph{functional compression}: realizing the target operation
through shared algebraic composition rather than rank reduction.}
We establish the following: 

\begin{itemize}[leftmargin=2em, labelsep=0.5em]

\item \textbf{Propagation of Local Alignment to Global Structure (Section~\ref{sec:decomposition}):}
We derive a Cauchy--Schwarz lower bound whose equality conditions require exact geometric alignment (collinearity) at every nonzero target entry.
These local alignment identities propagate through the target's support to enforce a globally shared balancedness condition and uniform rank.
{ Combining this global structure with the scalar AM--GM inequality yields an output-dependent lower bound on the collinear set, whose attainment requires all participating factor slices to be full-rank scaled unitaries with a globally shared norm.}

\item \textbf{Global Optimality and Universal Complexity Floor (Section~\ref{sec:optimality}):}
We derive a matrix AM--GM inequality extending our output-dependent lower bound globally to arbitrary parameters.
Under exact reconstruction, this establishes the universal complexity floor $\mathcal H_{\inf}(\delta)\geq 3 \, |\supp(\delta)|$, attained exactly at feasible factorizations with collinear, unitary slices on the target support.
The floor depends only on support size, providing a common baseline for comparing target complexity at a fixed order.
{ Targets lacking a floor-attaining factorization have constrained infima separated from the floor by a positive gap that is uniform at each fixed order.}

\item \textbf{Quasigroups: 
{ Functional Compression via Latent Homomorphism} 
(Section~\ref{sec:quasigroups}):}
For quasigroup targets (cancellative operations with no repeated
output within any row or column), floor attainment requires an
injective \textbf{homomorphism} into unitary matrices
(up to isotopy and unitary gauge), 
which enforces \emph{latent associativity} of the target operation. 
Consequently, the universal floor  is attained if and only if the target is isotopic to a group, 
and every global minimizer is equivalent to the underlying group's left-regular representation, formally resolving Conjecture~6.1 of \citet{huh2025discovering}.

{ 
\item \textbf{Beyond Quasigroups: Collision Rigidity and Reduced Unitary Factorization (Section~\ref{sec:beyond-quasigroups}):}
For non-quasigroup targets (operations with a repeated output within at least one row or column), 
floor attainment forces columns sharing an output within a row to be identical throughout,
 and likewise rows sharing an output in a column. 
Merging these duplicates yields a reduced table that embeds in a group-isotope table,
{ with composition realized by unitary matrix multiplication.}
Every floor-attaining factorization uses the fewest distinct input matrices compatible with exact reconstruction, 
keeps all participating matrices full-rank, and ensures that
each reduced input is uniquely determined by the output
and the other input (i.e., cancellative).

}

\item \textbf{Mechanized Verification:} 
All foundational theorems are formally verified in Lean~4.\footnote{\url{https://anonymous.4open.science/r/HyperCube_Lean-4664/README.md}} 
\end{itemize}

\section{Theoretical Framework}
\subsection{Algebraic Structure and Structure Tensor}
\label{sec:Algebraic_Structures}

Let $(Q, \circ)$ denote a finite set of $n$ elements equipped with a binary operation $\circ: Q \times Q \to Q$.
The operation table
is captured by its structure tensor $\delta \in \{0,1\}^{n \times n \times n}$, where $\delta_{abc} \coloneqq \mathbb{I}_{\{a \circ b = c\}}$ for $a,b,c \in Q$. { The \emph{support} of $\delta$ is $\supp(\delta)\coloneqq\{(a,b,c)\in Q^3 \mid \delta_{abc}=1\}$;}

A \emph{finite quasigroup} is a binary operation with the \emph{cancellative} property: i.e., the equations $a \circ x = c$ and $y \circ b = c$ have unique solutions for every $a, b, c \in Q$. { Equivalently, no output symbol repeats within any row or column of the operation table.}
A \emph{group} can be defined as a quasigroup that additionally satisfies 
\emph{associativity} $(a \circ b) \circ c = a \circ (b \circ c)$.

\subsection{Modeling Framework} 
\paragraph{Notations.}
For $X,Y \in \mathbb{C}^{n\times n}$, we define the { \textit{normalized} trace
$\Tr(X) \coloneqq \frac{1}{n}\text{Trace}(X)$,}
and the associated  \textit{normalized}
inner product and norm as  $\langle X,Y\rangle \coloneqq \Tr(X^\dagger Y)$ and $\|X\|^2 \coloneqq \langle X,X\rangle$. This normalization 
ensures $\|U\|^2 = 1$ for any unitary $U$. Throughout, $\dagger$ denotes the conjugate transpose.

\begin{figure}[t]
  \begin{center}
    \raisebox{0.275cm}{
  \includegraphics[pagebox=cropbox,width=0.225\textwidth]
  {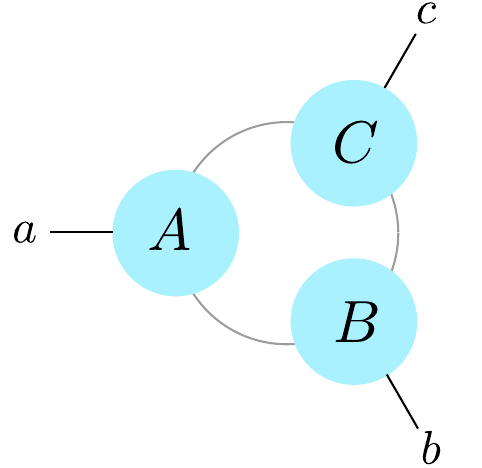}
}
  \qquad
  \includegraphics[pagebox=cropbox,width=0.25\textwidth]{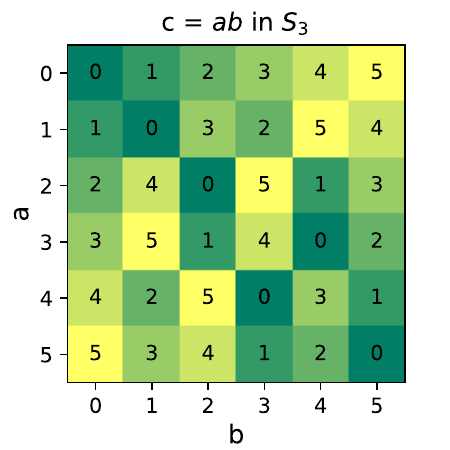}
  \qquad
  \includegraphics[pagebox=cropbox,width=0.25\textwidth]{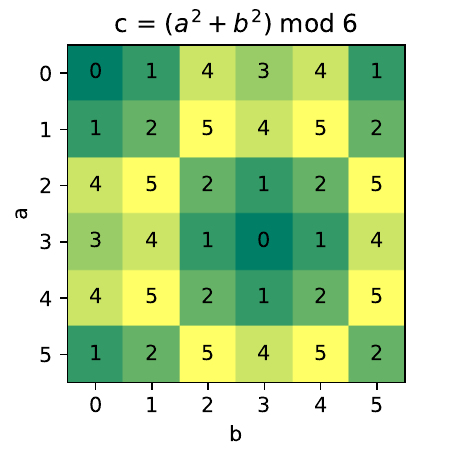}
\end{center}
  \vspace{-0.2cm}
    \caption{
    (Left) Illustration of the factorization model \eqref{eq:hypercube}         (Reproduced from \cite{huh2025discovering}).
    (Middle)  A group table ($S_3$).
    { (Right) A reducible, non-cancellative operation table ($(a^2+b^2) \bmod 6$).             }
    } 
    \label{fig:HyperCube}
\end{figure}

\paragraph{Operator-Valued Factorization.}
\citet{huh2025discovering}  defines a model
$T: Q^3 \to \mathbb{C}$ as a trilinear product of three factor tensors $\Theta \coloneqq (A,B,C) \in (\mathbb{C}^{n \times n \times n})^3$, illustrated in Figure~\ref{fig:HyperCube}:\footnote{Alternatively, this model can be interpreted as a mapping $T: Q^2 \to \mathbb{C}^n$ that produces a vector of logits over the output index $c$.}
\begin{equation}
\label{eq:hypercube} 
T_{abc}(\Theta) \coloneqq 
\Tr(A_a B_b C_c),
\end{equation}
where $A_a, B_b$, and $C_c$ denote the $n \times n$ matrix embeddings (linear operators) of $a,b,c \in Q$, referred to as factor slices.

\paragraph{{ Sensitivity} Objective.} 
\citet{huh2025discovering} introduce an objective that penalizes the model's \emph{ parameter sensitivity}---measured as the squared norms of the output gradients with respect to the parameters over supported triples (\emph{e.g.}, $\partial T_{abc}/\partial A_a = (B_bC_c)^\dagger$):
\begin{equation}\label{eq:H-Latin}
\mathcal H (\Theta)
{ ~~ = \sum_{(a,b,c)\in\supp(\delta)} n \, \| \nabla_\Theta T_{abc}(\Theta)\|^2 ~~ }
= \sum_{\supp(\delta)} \big( \|B_b C_c\|^2 + \| C_c A_a \|^2 + \| A_a B_b \|^2 \big)
\end{equation}
We study the objective's { constrained infimum} under exact target reconstruction, $T(\Theta)=\delta$:
\begin{align}
    \label{eq:constrained_opt}
    \mathcal H_{\inf}(\delta)
    =\inf_{\Theta\in\mathcal F_\delta}\mathcal H(\Theta).
\end{align}
We propose this value as a quantitative measure of \textbf{ algebraic complexity} for the target $\delta$.

\subsection{Formal Definitions}
\label{sec:definitions}

\begin{definition}[Feasible Set]
\label{def:feasible_set}
The feasible set $\mathcal F_\delta$ consists of  all parameters satisfying exact reconstruction:
$\mathcal F_\delta \coloneqq \{\Theta \mid T(\Theta)=\delta\}$.
{ By Proposition~\ref{prop:feasible-nonempty}, $\mathcal F_\delta$  is non-empty for any $\delta$.
We call $\Theta\in\mathcal F_\delta$ a factorization of $\delta$.}
\end{definition}

\begin{definition}[Unitary Set]
\label{def:unitary_set}
The unitary set $\mathcal{U}_\delta$ is the set of parameters $\Theta$ where every factor slice { on the support} is unitary,
$\mathcal{U}_\delta \coloneqq
\{\Theta \mid  A_a,\ B_b,\ C_{a\circ b}
\text{ \emph{are unitary for all} }a,b\in Q\}.$
\end{definition}

\begin{definition}[Isotopy]
\label{def:isotopy}
Two operation tensors $\delta$ and $\delta'$ are isotopic 
if they are equivalent up to independent permutations 
i.e.   $\exists (\phi, \psi, \chi) \in S_n^3$:
 such that $\delta'_{abc} = \delta_{\phi(a)\psi(b)\chi(c)}$.
\end{definition}

\subsection{Intrinsic Model Symmetries}
\label{sec:symmetries}

\paragraph{Gauge Invariance.}
The factorization model \eqref{eq:hypercube} is invariant under the continuous group of invertible \emph{gauge transformations}.
For any invertible matrices $U, V, W \in \text{GL}(n,\mathbb{C})$, the transformation
\begin{equation} \label{eq:gauge_transform}
\Theta' = (U A_a V^{-1}, \,  V B_b W^{-1}, \, W C_c U^{-1})
\end{equation}
preserves the product exactly:
$T_{abc}(\Theta') = T_{abc}(\Theta)$.
The objective \eqref{eq:H-Latin}
is invariant under \emph{unitary gauge transformations} only: i.e.
where $U, V, W$ are unitary matrices. 

\paragraph{Isotopy Equivariance.}
The factorization model \eqref{eq:hypercube} is equivariant under \emph{isotopy} (Definition \ref{def:isotopy}). For any bijections $(\phi, \psi, \chi) \in S_n^3$, permuting the parameters as $\Theta' = (A_{\phi(\cdot)}, B_{\psi(\cdot)}, C_{\chi(\cdot)})$ implies $T_{abc}(\Theta') = T_{\phi(a)\psi(b)\chi(c)}(\Theta)$. Consequently, the scalar objective \eqref{eq:H-Latin} is \emph{isotopy-invariant}. This symmetry greatly simplifies the subsequent analysis by allowing us to reduce any quasigroup target to its loop isotope (a quasigroup with a designated identity $e$) without loss of generality.

\section{Orthogonal Decomposition and Geometric Alignment}
\label{sec:decomposition}

Minimizing the quartic objective \eqref{eq:H-Latin} subject to the trilinear constraints $T(\Theta)=\delta$ yields a highly non-convex optimization problem \eqref{eq:constrained_opt}.
Direct stationarity analysis yields highly coupled polynomial tensor equations.
We instead pursue a global bounding strategy by first establishing a Cauchy--Schwarz lower bound. We prove that attaining this bound requires exact geometric alignment (collinearity), then characterize the algebraic consequences of this alignment.
We assume \emph{$T_{abc} \neq 0$ for all supported triples throughout this section}.

\begin{lemma}[Cauchy--Schwarz Bound] 
\label{lem:decomposition}
For any target $\delta$ and any parameter $\Theta$ with non-vanishing { participating} slices, the { sensitivity} objective \eqref{eq:H-Latin} is bounded below by
\begin{equation}
    \label{eq:B}
    \mathcal{H}(\Theta) \geq \mathcal{B}_{\delta}(\Theta) 
    \coloneqq \sum_{\supp(\delta)} |T_{abc}(\Theta)|^2\left(\frac{1}{\|A_a\|^2}+\frac{1}{\|B_b\|^2}+\frac{1}{\|C_c\|^2}\right), 
\end{equation}
where $\mathcal{B}_{\delta}$ acts as an inverse $\ell_2$ penalty at fixed model output.
 This bound is attained \emph{if and only if} the factors satisfy the 
\textbf{collinear identities} for every supported triple:
\begin{equation}
    \label{eq:alignment}
    B_b C_c = T_{abc} \frac{A_a^\dagger}{\|A_a\|^2}, \quad C_c A_a = T_{abc} \frac{B_b^\dagger}{\|B_b\|^2}, \quad A_a B_b = T_{abc} \frac{C_c^\dagger}{\|C_c\|^2}.
\end{equation}
\end{lemma}

\begin{proof}[Proof Sketch]
By the Cauchy--Schwarz inequality, the inner product $T_{abc} = \langle A_a^\dagger, B_b C_c \rangle$ yields the lower bound $\|B_b C_c\|^2 \ge |T_{abc}|^2 / \|A_a\|^2$. Summing this symmetric bound across all three gradient terms and over the target support directly establishes $\mathcal{H}(\Theta) \ge \mathcal{B}_{\delta}(\Theta)$.
Equality holds \emph{if and only if} the Cauchy--Schwarz bound saturates for every supported triple, which requires the matrices in each inner product to be collinearly aligned, yielding exactly \eqref{eq:alignment}. The full proof is in Appendix~\ref{app:deferred_proofs}.
\end{proof}

\begin{definition}[Collinear Set] 
\label{def:collinear_manifold}
The collinear set $\mathcal{M}_\delta$ consists of parameters that attain the Cauchy--Schwarz bound \eqref{eq:B}:
    $\mathcal{M}_\delta \coloneqq \{ \Theta \mid \mathcal{H}_{\delta}(\Theta) = \mathcal{B}_{\delta}(\Theta)  \}.$
By Lemma~\ref{lem:decomposition}, $\Theta \in \mathcal{M}_\delta$ if and only if it satisfies \eqref{eq:alignment} for every supported triple.
\end{definition}

{  
Minimizing sensitivity requires minimizing the sum of the lower bound $\mathcal{B}_\delta(\Theta)$ and the nonnegative alignment gap $\mathcal{H}(\Theta) - \mathcal{B}_\delta(\Theta)$. 
We first analyze the case where this gap vanishes by restricting to the collinear set $\mathcal{M}_\delta$.
The next two lemmas show that the \emph{ local} collinear identities \eqref{eq:alignment} propagate across the target's support, forcing a 
\emph{ shared global structure}.
}

\begin{lemma}[Shared Balancedness\footnote{ Analogous to balancedness between adjacent layers in deep linear networks
\citep{
    aroraImplicitRegularizationDeep2019a,huh2020curvature},
collinearity here forces equality of normalized Gram matrices.}]
\label{lem:index-independent-gram}
For { any target} $\delta$ and any 
{ collinear} 
$\Theta \in \mathcal{M}_\delta$,
every supported triple $(a,b,c)$ admits normalized positive semi-definite Gram matrices 
$X,Y,Z\in\mathbb C^{n\times n}$ satisfying
\begin{equation}
    \label{eq:gram-identities}
    X = \frac{C_c C_c^\dagger}{\|C_c\|^2} = \frac{B_b^\dagger B_b}{\|B_b\|^2}, \quad 
    Y = \frac{A_a A_a^\dagger}{\|A_a\|^2} = \frac{C_c^\dagger C_c}{\|C_c\|^2}, \quad 
    Z = \frac{B_b B_b^\dagger}{\|B_b\|^2} = \frac{A_a^\dagger A_a}{\|A_a\|^2}.
\end{equation}
The trace of these matrices is $\Tr(X) = \Tr(Y) = \Tr(Z) =  1$.
For fixed $\Theta$,  the matrix $Z$ satisfying \eqref{eq:gram-identities} is \emph{ globally shared} across all supported triples.
For quasigroup targets, $X$ and $Y$ are also globally shared.
The proof is provided in Appendix~\ref{app:deferred_proofs}.
\end{lemma}

{
The next lemma shows that $X, Y,$ and $Z$ become orthogonal projections when scaled by a common factor $\kappa$, 
identified as their normalized rank. Although defined as a continuous ratio between the factor-norm product and the output magnitude, 
$\kappa$ is confined to discrete values $\{1/n, 2/n, \ldots, 1\}$.
}

\begin{lemma}[Normalized Rank $\kappa$]
\label{lem:proj-kappa}
For { any target} $\delta$ and { collinear} $\Theta \in \mathcal{M}_\delta$,
the dimensionless ratio $\kappa_{abc} \coloneqq \|A_a\|^2\|B_b\|^2\|C_c\|^2 / |T_{abc}|^2$ is constant across the support. This constant $\kappa$ enforces the strict projection identity  $Z = \kappa Z^2$, thereby identifying $\kappa$ as the \emph{normalized rank}:
\begin{equation}
    n \kappa = {\operatorname{rank}(Z)}
    = {\operatorname{rank}(A_a)} = {\operatorname{rank}(B_b)} = {\operatorname{rank}(C_c)}
    \le n .
\end{equation}
Thus, $\kappa \leq 1$.
Moreover, $\kappa=1$ \emph{if and only if} 
$X=Y=Z=I_n$ { on the support}.
\end{lemma}
\begin{proof}[Proof Sketch]
Substituting the collinear identities \eqref{eq:alignment} into the shared matrix $Z$ yields $Z=\kappa_{abc}Z^2$. 
Since $Z$ is globally shared and nonzero,
$\kappa_{abc}$ must be a constant $\kappa$. 
Defining $P \coloneqq \kappa Z$ yields $P^2 = \kappa^2 Z^2 = \kappa Z = P$, so $P$ is an orthogonal projection with eigenvalues in $\{0,1\}$.
Thus, $\operatorname{rank}(Z)
= 
\operatorname{rank}(P) =
n\Tr(P) = n \kappa \Tr(Z) = n \kappa$.
Eq~\eqref{eq:gram-identities}  then gives rank $n\kappa$ to every participating slice. The same identity holds for 
$X,Y$; at $\kappa=1$, $X,Y,Z$ are \emph{rank-n} projections and hence equal $I_n$.
\end{proof}

{

\begin{remark}
Unlike standard norm-product estimates $|T_{abc}|^2\leq n\|A_a\|^2\|B_b\|^2\|C_c\|^2$, Lemma~\ref{lem:proj-kappa} establishes the upper bound $\|A_a\|^2\|B_b\|^2\|C_c\|^2\leq |T_{abc}|^2$ on the collinear set across the target's support. 
Together with scalar AM--GM, this inequality yields a sensitivity lower bound that depends on the parameters only through the model output 
(Lemma~\ref{lem:AMGM}).

\end{remark}
}

\begin{lemma}[ Collinear Output-Dependent Bound and Inverse-Rank Penalty]
    \label{lem:AMGM} For { any target} $\delta$ and any { collinear} $\Theta \in \mathcal{M}_\delta$,
the sensitivity objective satisfies the following bounds:
\begin{align}
        \mathcal{H}(\Theta)  
                 \, &  \ge \, 
          \kappa^{-1/3}  \, \mathcal{H}^*_{\delta}(T(\Theta))  
        \, \ge  \,        \mathcal{H}^*_{\delta}(T(\Theta)) 
    \, \eqqcolon \,
    3 \,  \sum_{\supp(\delta)} |T_{abc}|^{4/3} 
    \label{eq:collinear_lower_bound}
\end{align}
Equality holds in the first bound if and only if the norms of all { participating}
factor slices are \emph{ globally} equal
($\|A_a\|^2=\|B_b\|^2=\|C_c\|^2$ on the support), and in the second  if and only if $\kappa=1$.
\end{lemma}
\begin{proof}[Proof Sketch.]
    For $\Theta \in \mathcal{M}_\delta$, 
        $\mathcal{H}(\Theta) = \mathcal{B}_{\delta}(\Theta)$.
    For a supported triple $(a,b,c)$, let
    $\alpha_a \coloneqq \|A_a\|^{-2}$,
    $\beta_b \coloneqq \|B_b\|^{-2}$, and
    $\gamma_c \coloneqq \|C_c\|^{-2}$.
    Note that $\alpha_a \beta_b \gamma_c = \kappa^{-1} |T_{abc}|^{-2}$ 
        (Lemma~\ref{lem:proj-kappa}).
        Applying the {\emph{scalar} AM--GM} inequality ($\alpha+\beta+\gamma \ge 3(\alpha\beta\gamma)^{1/3}$) to the definition of $\mathcal{B}_{\delta}(\Theta)$ 
    \eqref{eq:B}
    yields the first inequality,     which saturates \emph{if and only if} 
        $\alpha_a = \beta_b= \gamma_c$ 
        for every supported triple $(a,b,c)$. 
            The second inequality follows from $\kappa\leq1$, with equality \emph{if and only if}  $\kappa=1$,
    { since $\mathcal H_\delta^*(T(\Theta))>0$}.
\end{proof}

{
}

{
These results show how local alignment identities propagate across the target support: the shared normalized Gram matrix $Z$ is a scaled orthogonal projection, and all participating factor slices have rank $n\kappa$.
This shared global structure underpins the output-dependent lower bound $\mathcal{H}(\Theta) \ge \mathcal{H}^*_\delta(T(\Theta))$. Attaining this bound within $\mathcal{M}_\delta$ further requires all participating slices to be scaled unitaries with a globally shared scale $\alpha>0$.
Thus, $\Theta\in\mathcal{M}_\delta\cap\alpha\mathcal{U}_\delta$, where
$\alpha\mathcal{U}_\delta\coloneqq\{\alpha\Theta':\Theta'\in\mathcal{U}_\delta\}$.
The empirical results show that this bound holds even outside that set (Figure~\ref{fig:associativity_violation}, Left, Middle). 
We next extend the output-dependent bound from the collinear set to arbitrary parameters.
}

\section{Global Lower Bounds and Universal Complexity Floor}
\label{sec:optimality}

{
In this section, 
we derive the \emph{matrix AM--GM inequality} (Lemma~\ref{lem:matrix_AMGM}) to extend the output-dependent lower bound of Lemma~\ref{lem:AMGM} to arbitrary parameters (Theorem~\ref{thm:unconditional_bound}).
Under exact reconstruction $T(\Theta)=\delta$, this yields
the universal complexity floor $\mathcal H_{\inf}(\delta) \ge \mathcal H^*_\delta(\delta)=3 \, |\supp(\delta)|$ (Theorem~\ref{thm:absolute_lower_bound}). The floor depends only on the target's support size $|\supp(\delta)|=n^2$.
Theorem~\ref{thm:positive-infimum-gap} shows that, if the floor is unattained, the constrained infimum is separated from it by a positive gap that is uniform at each fixed order.
}

\begin{lemma}[Matrix AM--GM]
\label{lem:matrix_AMGM}

For all $X,Y,Z \in \mathbb{C}^{n \times n}$,
let $\theta \coloneqq (X,Y,Z)$
and $t(\theta)\coloneqq\Tr(XYZ)$. Then
\begin{equation}
    { n \, \|\nabla_\theta \, t\|^2}
    = \|XY\|^2+\|YZ\|^2+\|ZX\|^2
    \ge 3 \,|t|^{4/3}.
    \label{eq:matrix_amgm}
\end{equation}
{For $t  \neq 0$,}
equality holds if and only if
$X,Y,Z$ are scaled unitaries 
that satisfy \[
\frac{XX^\dagger}{\|X\|^2}  = \frac{YY^\dagger}{\|Y\|^2} = \frac{ZZ^\dagger}{\|Z\|^2}=I_n,
\qquad \|X\|^2=\|Y\|^2=\|Z\|^2,
\qquad XYZ=tI_n,
\] 
implying  $\kappa \coloneqq \|X\|^2\|Y\|^2\|Z\|^2/|t|^2=1$.
{Equality condition for $t=0$ is $XY=YZ=ZX=0$.
}

\end{lemma}

\begin{proof}[Proof Sketch]
Define a block-cyclic matrix $M \in\mathbb{C}^{3n\times 3n}$ and evaluate its products:
\[
M \coloneqq
\begin{pmatrix}
0 & X & 0 \\
0 & 0 & Y \\
Z & 0 & 0
\end{pmatrix},
\qquad
M^2=
\begin{pmatrix}
0 & 0 & XY \\
YZ & 0 & 0 \\
0 & ZX & 0
\end{pmatrix},
\qquad
M^3=
\begin{pmatrix}
XYZ & 0 & 0 \\
0 & YZX & 0 \\
0 & 0 & ZXY
\end{pmatrix}.
\]
Evaluating the normalized norm of $M^2$ and the normalized trace of $M^3$  yields (via trace cyclicity): \[
\Tr\left({M^2}^\dagger M^2\right) = \frac{1}{3}\left(\|XY\|^2+\|YZ\|^2+\|ZX\|^2\right),
\qquad 
\Tr(M^3) = \Tr(XYZ).
\]
Applying the Spectral Trace-Frobenius Bound 
$ \Tr({M^2}^\dagger M^2)
    \ge     \big | \Tr(M^3) \, \big |^{4/3}$
(Lemma~\ref{lem:spectral_trace_frob}  in Appendix~\ref{app:matrix-AM-GM})
yields the inequality \eqref{eq:matrix_amgm} and its equality condition.
\end{proof}

\begin{theorem}[ Global Output-Dependent Bound]
\label{thm:unconditional_bound}
For { any target} $\delta$, the output-dependent lower bound $\mathcal H^*_\delta(T(\Theta))$ in Lemma~\ref{lem:AMGM} holds for arbitrary parameters $\Theta$, without assuming collinearity:
\begin{equation}
    \label{eq:unconditional_lower_bound}
    \qquad \mathcal{H}(\Theta) \ge \mathcal{H}^*_{\delta}(T(\Theta)) 
    {
    \, \coloneqq  3 \,  \sum_{\supp(\delta)}
    |T_{abc}|^{4/3}.}
\end{equation}
Equality holds if and only if 
the equality conditions of Lemma~\ref{lem:matrix_AMGM} hold \emph{ globally} for every supported triple.
\end{theorem}

\begin{theorem}[ Feasible-Set Lower Bound] \label{thm:absolute_lower_bound}
For { any target} $\delta$ and any feasible factorization $\Theta \in \mathcal{F}_\delta$, the sensitivity objective is bounded below by $\mathcal H^*_\delta(\delta)$: \begin{equation}
    \label{eq:absolute_lower_bound}
    \mathcal{H}(\Theta) \ge \mathcal{H}_{\inf}(\delta) \ge 
    \mathcal H^*_\delta(\delta).
\end{equation}    
Equality holds throughout \emph{if and only if} 
all factor slices { on the support} are collinear and unitary:
$\Theta\in\mathcal F_\delta\cap\mathcal M_\delta\cap\mathcal{U}_\delta$.
\end{theorem}

\begin{proof}[Proof of Theorems~\ref{thm:unconditional_bound} and \ref{thm:absolute_lower_bound}]
Lemma~\ref{lem:matrix_AMGM} establishes a local lower bound for any triple $(a,b,c)$: 
\begin{equation} \label{eq:local_amgm}
    { n \, \|\nabla_\Theta \, T_{abc} \|^2 \, = \, }
    \|A_a B_b\|^2 + \|B_b C_c\|^2 + \|C_c A_a\|^2 \, \ge \, 3 \, |T_{abc}|^{4/3}.    
\end{equation}
Summing \eqref{eq:local_amgm} over $\supp(\delta)$ yields the unconditional global lower bound \eqref{eq:unconditional_lower_bound}.
Equality holds \emph{if and only if} \eqref{eq:local_amgm} saturates for every supported triple (where $\delta_{abc}=1$). 
Restricting to $\mathcal{F}_\delta$ (i.e., $T(\Theta)=\delta$) yields $\mathcal{H}^*_{\delta}(\delta)$. 
On the feasible set, $T_{abc}=1$ for each supported triple, 
{under which} the equality conditions of
Lemma~\ref{lem:matrix_AMGM} yield { slice-norm equality and $\kappa=1$, so the { participating} slices are unitary and $A_aB_bC_c=I_n$, which imply collinearity \eqref{eq:alignment} on every supported triple.}
Conversely, unitary collinear slices make each supported local sensitivity equal to $3$, attaining the floor.
\end{proof}

\vspace{-1.0em}
\paragraph{Universal Complexity Floor:}
The complexity floor $\mathcal H^*_{\delta}(\delta)$ depends only on the target's support size
$|\supp(\delta)|=n^2$,
not on its algebraic structure:
since $\delta_{abc}=1$ on $\supp(\delta)$,
\begin{equation}
\label{eq:universal_baseline}
    \mathcal H^*_{\delta}(\delta)  
        = 3 \, \sum_{\supp(\delta)}  1  = 3 \, |\supp(\delta)|  
    = 3 \, n^2. 
\end{equation}
Thus, all targets of the same order share a universal complexity floor, { which allows comparing their constrained infima $\mathcal H_{\inf}(\delta)$ through their excess above this floor.}

{
\begin{theorem}[Positive Infimum Gap] \label{thm:positive-infimum-gap}
For every fixed $n\geq1$, there exists $\varepsilon_n>0$, 
such that every binary-operation target $\delta$ of 
order $n$ 
either admits a feasible factorization $\Theta_*\in\mathcal F_\delta$ 
that attains the floor
$\mathcal H(\Theta_*)= 3  \, |\supp(\delta)|$, or satisfies
\begin{equation}
\label{eq:positive_gap}
\mathcal H(\Theta)\geq\mathcal H_{\inf}(\delta) \geq 
3  \, |\supp(\delta)|  + \varepsilon_{n}
\qquad \text{for all }\Theta\in\mathcal F_\delta.
\end{equation}
\end{theorem}

In particular, a constrained infimum equal to the floor is attained. Lemma~\ref{lem:spectral-no-escape} supplies the key compactness step by bounding feasible sequences approaching the floor after unused output slices are set to zero. The proof is given in Appendix~\ref{app:positive_gap}.
}

\section{Quasigroups: Group Isotopy  Minimizes Algebraic Complexity}
\label{sec:quasigroups}

{ 
We now characterize the algebraic structure of $\delta$ required to attain this complexity floor.
Among quasigroup targets, Theorem~\ref{thm:unitary_equivalence} establishes that any feasible unitary collinear factorization must realize an \emph{injective homomorphism}, forcing the target to be a \emph{group isotope}.
Furthermore, every global minimizer is equivalent to the underlying group's unitary left-regular representation, 
resolving the central conjecture of \citet{huh2025discovering}.
Consequently, the complexity floor is attained exactly for group-isotope targets 
(Theorem~\ref{thm:global_optimality_quasigroup}). 
}

\begin{theorem}[Unitary Collinearity $\iff$ Group Isotope]
\label{thm:unitary_equivalence}
For any finite quasigroup target $\delta$, the unitary feasible collinear set ($\mathcal{F}_\delta \cap \mathcal{M}_\delta \cap \mathcal{U}_\delta$) is non-empty \emph{if and only if} $\delta$ is isotopic to a group.
\end{theorem}
\begin{proof}[Proof Sketch]
(\emph{Necessity} $\implies$): By isotopy equivariance, we can assume without loss of generality that $\delta$ is a \emph{loop} (a quasigroup with an identity element $e$). A unitary collinear factorization admits a unitary gauge transformation 
\eqref{eq:gauge_transform}
that synchronizes the matrix embeddings: $A'_g = B'_g = (C'_g)^\dagger \coloneqq \rho(g)$ (Lemma~\ref{lem:app_synchronization}).
{ 
Then, the collinear identities \eqref{eq:alignment} and feasibility ($T=\delta$) force $\rho$ to be an \textbf{injective homomorphism} (Lemma~\ref{lem:app_homomorphism}). 
That is, $\rho$ must embed the operation $\circ$ into matrix multiplication:
\begin{equation}
    \rho(a\circ b) = \rho(a)\rho(b).
\end{equation}
}Consequently, the { loop target's} operation $\circ$ {must inherit} the \textbf{associativity} of matrix multiplication:
$\rho((x \circ y) \circ z) = \rho(x)\rho(y)\rho(z) = \rho(x \circ (y \circ z))$. \\
(\emph{Sufficiency} $\impliedby$): If $\delta$ is a group isotope, its left-regular representation defines a unitary collinear factorization that achieves $T=\delta$
(Lemma~\ref{lem:app_group_existence}).
(See Appendix~\ref{app:unitary_proof} for the full proof).
\end{proof}

\begin{theorem}[Global Optimality for Quasigroups]
\label{thm:global_optimality_quasigroup}
For any quasigroup target $\delta$, the  constrained infimum satisfies $\mathcal H_{\inf}(\delta) \geq 3 \, |\supp(\delta)|$.
This universal floor is attained if and only if $\delta$ is isotopic to a group.
In that case, every global minimizer is equivalent to the left-regular representation of the underlying group, up to isotopy and unitary gauges, as described in Lemma~\ref{lem:app_representation_uniqueness}.
{ Otherwise, the feasible infimum is separated from the universal floor by a uniform associativity gap \eqref{eq:positive_gap}.}
\end{theorem}
\begin{proof}
On $\mathcal F_\delta$, Theorem~\ref{thm:absolute_lower_bound} shows that the universal floor is attained exactly at $\mathcal F_\delta\cap\mathcal M_\delta\cap\mathcal{U}_\delta$.
Theorem~\ref{thm:unitary_equivalence} shows that this intersection is non-empty if and only if $\delta$ is isotopic to a group;
in this case, Lemma~\ref{lem:app_representation_uniqueness} classifies every global minimizer up to the stated equivalences.
Otherwise, the floor is unattained, and { Theorem~\ref{thm:positive-infimum-gap} gives the uniform positive infimum gap.}
\end{proof}

{
}

\subsection{Empirical Analysis: Learning Dynamics and the Associativity Gap}
\label{sec:empirical}

\begin{figure*}[t]
  \vskip -0.1in
  \begin{center}
    \includegraphics[width=0.995\textwidth]{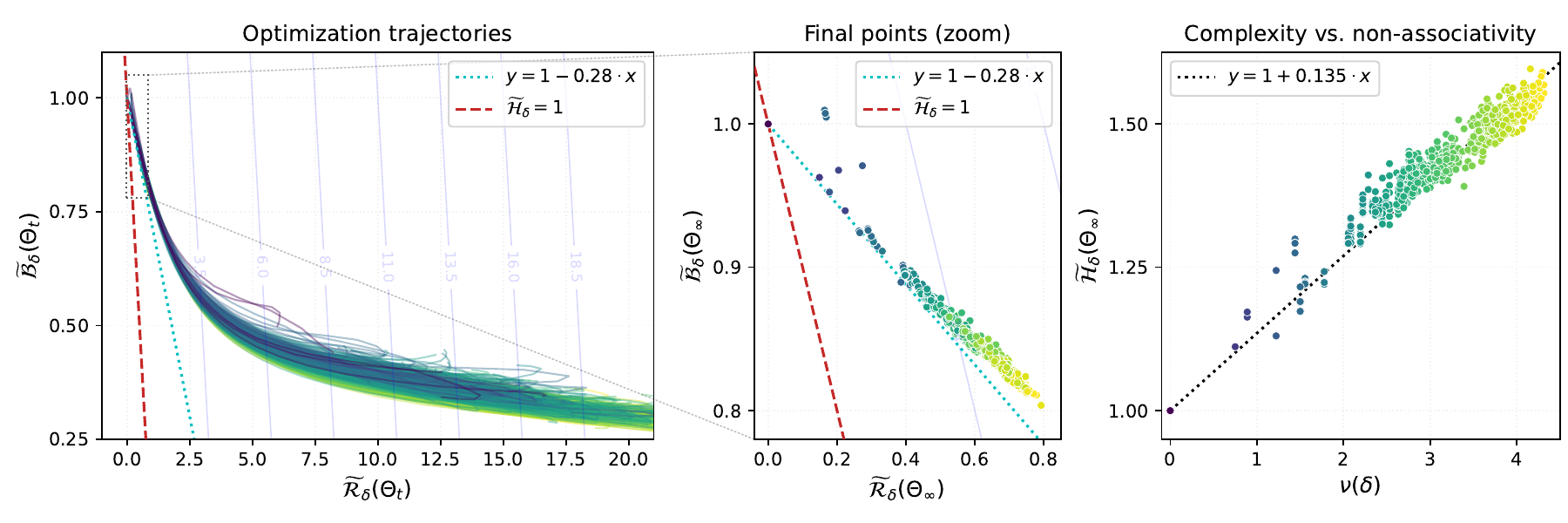}
  \vskip -0.1in
    \caption{
    \textbf{Learning dynamics and attained objective values.}
    \textbf{Left:} Optimization trajectories $\Theta_t$ in the $(\widetilde{\mathcal R}_\delta, \widetilde{\mathcal B}_\delta)$ plane from random initialization.
    \textbf{Middle:} Final objective components for each target, { enlarged from the dotted box on the left}, with training endpoints denoted by $\Theta_\infty$.
    \textbf{Right:} Normalized complexity $\widetilde{\mathcal H}_\delta(\Theta_\infty)$ against normalized latent associativity violation $\nu(\delta)$
    \eqref{eq:associativity_violation}.
    Here, $\mathcal R_\delta \coloneqq \mathcal H - \mathcal B_\delta$ is the alignment gap, and tildes denote division by the output-dependent lower bound $\mathcal H_\delta^*(T(\Theta))$.
    Colors indicate $\nu(\delta)$, and contours show constant values of $\widetilde{\mathcal H}_\delta$.
    { The red contour marks $\widetilde{\mathcal H}_\delta=1$, the lower bound of Lemma~\ref{lem:AMGM}, established globally by Theorem~\ref{thm:unconditional_bound}.
    }
    }
  \label{fig:associativity_violation}
  \end{center}
  \vskip -0.2in
\end{figure*}

We confirm our theoretical findings via numerical optimization of the regularized objective $\mathcal L(\Theta)=\|T(\Theta)-\delta\|_F^2 +  \lambda \mathcal H(\Theta)$. 
We evaluate targets of all loop isomorphism classes of orders $5$ and $6$, alongside random samples of $100$ loops each at orders $7$ and $8$, optimizing via full-batch gradient descent.
This formulation initially relaxes the exact feasibility constraint ($\lambda=0.05$); in the final phase, we set $\lambda=0$ to fit the target exactly (see Appendices~\ref{app:experiments_hyperparameters} and~\ref{app:population_results}).

We track the sensitivity and the Cauchy--Schwarz bound, both normalized by the output-dependent bound: 
$\widetilde{\mathcal H}_\delta \coloneqq \mathcal H/\mathcal H^*_\delta(T(\Theta))$ and 
$\widetilde{\mathcal B}_\delta \coloneqq \mathcal B_\delta/\mathcal H^*_\delta(T(\Theta))$.
Their difference is the normalized alignment gap, 
$\widetilde{\mathcal R}_\delta \coloneqq \widetilde{\mathcal H}_\delta - \widetilde{\mathcal B}_\delta$.
In Figure~\ref{fig:associativity_violation}, the left panel shows full optimization trajectories $(\widetilde{\mathcal R}_\delta(\Theta_t), \widetilde{\mathcal B}_\delta(\Theta_t))$, and the middle panel shows the training endpoints. 
{ 
The trajectories start in the lower right with a large alignment gap $\widetilde{\mathcal R}_\delta$ and sweep smoothly \emph{right-to-left} toward smaller $\widetilde{\mathcal R}_\delta$ and larger $\widetilde{\mathcal B}_\delta$, 
decreasing $\widetilde{\mathcal H}_\delta$ (shown as contour lines). 
Figure~\ref{fig:detailed_dynamics} in Appendix~\ref{app:experiments} illustrates sample trajectories over training steps, showing smooth optimization dynamics without visible stagnation for one group and one non-group target.

The fact that all points lie on the right side of the red contour line is consistent with the global output-dependent lower bound ($\widetilde{\mathcal{H}}_\delta \ge 1$) established by Theorem~\ref{thm:unconditional_bound}.
The cyan dotted reference suggests a stronger, alignment-dependent lower bound: the observed normalized excess complexity $\widetilde{\mathcal H}_\delta-1$ grows with the normalized alignment gap.
}

\paragraph{Training Endpoints and the Associativity Gap.}
The training endpoints are consistent with the floor-attainment characterization and positive infimum gap of Theorem~\ref{thm:global_optimality_quasigroup} (Figure~\ref{fig:associativity_violation}, middle and right).
{ We measure latent associativity violation $\nu(\delta)$ by the minimum normalized count of associativity failures over reduced loop isotopes.}\footnote{
Because the objective $\mathcal{H}_{\inf}$ is invariant under isotopy, we evaluate targets based on their most associative isotopic representative. Let $\mathfrak{L}(\delta)$ denote the finite set of reduced loop tables isotopic to a target $\delta$. We define the \emph{normalized latent associativity violation count} as:
\begin{equation}
    \label{eq:associativity_violation}
    \nu(\delta) \coloneqq \min_{(Q,\star)\in\mathfrak{L}(\delta)} \frac{1}{|\supp(\delta)|} \sum_{a,b,c\in Q} \mathbb{I}_{\{(a\star b)\star c \neq a\star(b\star c)\}}.
\end{equation}
This metric is \emph{isotopy-invariant} and vanishes if and only if $\delta$ is { isotopic to a group.}
}
For group-isotope targets ($\nu(\delta)=0$), the training endpoints approach the complexity floor $\widetilde{\mathcal{H}}_\delta = 1$ at $(\widetilde{\mathcal R}_\delta,\widetilde{\mathcal B}_\delta)=(0,1)$.
For targets not isotopic to a group, the training endpoints retain a positive alignment gap and complexity above the universal floor.
The theory excludes feasible collinear factorizations for these targets and establishes a { positive infimum gap}.
Across targets, larger $\nu(\delta)$ is associated with both larger normalized alignment gaps and higher normalized complexity, connecting the middle and right panels.
This empirical association suggests a graded relationship beyond the theorem's zero-versus-positive gap characterization.

{ 
\section{Beyond Quasigroups: Collision Rigidity and Reduced Unitary Factorization}
\label{sec:beyond-quasigroups}

We extend the analysis to arbitrary binary operations beyond quasigroups, allowing failures of cancellation. 
We retain full tensor feasibility, including zero predictions on unsupported triples.

\paragraph{Collision rigidity.}
A \emph{collision} is a repeated output within a row or column at distinct input labels, hence a failure of cancellation (e.g., $a\circ b=a\circ b'$ with $b\ne b'$). 
By the equality case of Theorem~\ref{thm:absolute_lower_bound}, a feasible floor-attaining factorization has unitary participating slices and satisfies $A_a B_b = C_{a\circ b}^\dagger$.
For such a row collision, canceling the unitary $A_a$ gives $B_b=B_{b'}$. 
This matrix equality implies 
$T_{xbc}=T_{xb'c}$ for every $x,c$; 
full one-hot feasibility then forces $x\circ b=x\circ b'$ for every $x$. Thus, a collision in one row forces identical entire columns; 
likewise, a collision in one column forces identical entire rows.
If all rows and columns are distinct, floor attainment forces the target to be a quasigroup; Theorem~\ref{thm:global_optimality_quasigroup} then requires group isotopy.
For arbitrary operations, the necessary condition is
\begin{equation}
\label{eq:general-floor-structure}
\mathcal{H}(\Theta)=3 \, |\supp(\delta)|,\quad\Theta\in\mathcal{F}_\delta
\quad\Longrightarrow\quad
\text{group isotope, or repeated rows, or repeated columns}.
\end{equation}
Redundancy alone does not guarantee attainment. Detailed proofs appear in Appendix~\ref{app:beyond-quasigroups}.

\paragraph{Reduced table and exact factorization.}
Identifying identical rows ($a\sim_R a'$) and identical columns ($b\sim_C b'$) yields a reduced table $\bar\circ:R_\circ\times S_\circ\to U_\circ$, where $R_\circ=Q/{\sim_R}$, $S_\circ=Q/{\sim_C}$, and $U_\circ$ is the set of used outputs. 
The sets $R_\circ$, $S_\circ$, and $U_\circ$ can have different cardinalities, so the reduced table need not be a quasigroup table.
The following characterization is proved in Appendix~\ref{app:beyond-quasigroups}.

\begin{theorem}[Exact Reduced Unitary Factorization]
\label{thm:exact-reduced-factorization}
For a binary operation on $Q$ with $|Q|=n\geq1$, there exists $\Theta\in\mathcal F_\delta$ with $\mathcal H(\Theta)=3 \, |\supp(\delta)|$ if and only if the reduced table admits unitary maps $U:R_\circ\to\mathrm U(n)$, $V:S_\circ\to\mathrm U(n)$, and $W:U_\circ\to\mathrm U(n)$ satisfying

$$U_rV_s=W_{r\mathbin{\bar\circ}s}^\dagger, \qquad \Tr(W_c^\dagger W_d)=\mathbb I_{\{c=d\}} \quad(r\in R_\circ,\ s\in S_\circ,\ c,d\in U_\circ),$$

Any such maps $U,V,W$ are necessarily injective, and every feasible floor point descends to them via $A_a=U_{[a]_R}$, $B_b=V_{[b]_C}$, and $C_c=W_c$ for $c\in U_\circ$.
\end{theorem}

\paragraph{Input compression and conditional recoverability.}
Exact reconstruction requires distinct matrices $A_a$ for distinct rows, and distinct matrices $B_b$ for distinct columns.
At the floor, the full-rank unitary input matrices attain the smallest numbers of distinct matrices compatible with exact reconstruction:\footnote{More generally, any separate deterministic encoders $e_R:Q\to E_R$ and $e_C:Q\to E_C$ into finite alphabets admitting a decoder $D$ with $D(e_R(a),e_C(b))=a\circ b$ for every pair satisfy $|R_\circ|\leq|E_R|$ and $|S_\circ|\leq|E_C|$. The row and column quotient encoders attain both bounds, independently of floor attainment.}
\[
|\{A_a:a\in Q\}|=|R_\circ|, \qquad |\{B_b:b\in Q\}|=|S_\circ|.
\]
At the floor, the reduced table is injective in each input separately (Theorem~\ref{thm:general-reduced-table}), 
so either reduced input is recoverable given the output and the other input. This conditional recoverability alone is weaker than floor attainment: 
every quasigroup satisfies this property.

The unitary product identity also shows that the reduced table embeds in a group-isotope table: after independently relabeling its rows, columns, and outputs, it occurs as a subtable of group multiplication. For general reduced tables, this embeddability is necessary but not sufficient for floor attainment, which requires the prescribed dimension $n$ and orthonormal output matrices
(see the three-symbol parity example in Appendix~\ref{app:finite-group-realizations}). Together with the product identity, this orthonormality ensures exact reconstruction, with unused output slices set to zero.

\paragraph{Stray collisions and the positive complexity gap.}
A collision is \emph{stray} if two inputs share an output within a row or column, but their corresponding entire columns or rows are not identical. Such a collision contradicts cancellation in the reduced table. By Corollary~\ref{cor:general-stray-collision} and Theorem~\ref{thm:positive-infimum-gap}, any operation with a stray collision cannot attain the floor and has a positive infimum gap above $3 \, |\supp(\delta)|$. Appendix~\ref{app:extension-examples} illustrates this obstruction with the associative operation $a\circ b=\min(a,b)$ on $\{0,1\}$, and also exhibits a redundant floor-attaining operation with no associative isotope.
}

\section{Conclusion}
\label{sec:conclusion}

We characterize the \emph{ algebraic complexity} of finite binary operations, $\mathcal H_{\inf}(\delta)$, the constrained infimum of the sensitivity objective of the canonical tensor-factorization model.
This complexity
favors \emph{ functional compression}: at the universal floor, all participating factor slices are full rank, and simplicity arises from reusing shared symbol representations whose matrix multiplication directly implements the target operation ($A_aB_b=C_{a\circ b}^{\dagger}$).
This contrasts with  low-rank matrix completion, which exploits \emph{ geometric compression} through shared linear dependencies.

Attaining this universal complexity floor requires local alignment identities to propagate consistently across the operation table.
For quasigroups, floor attainment characterizes group isotopes: every global minimizer realizes the underlying group's left-regular representation---an \emph{injective homomorphism}---up to isotopy and unitary gauge.
This characterization resolves the central conjecture of \citet{huh2025discovering}.
{ Beyond quasigroups, floor attainment enforces collision rigidity: a collision within a row forces the corresponding columns to be identical, and a collision within a column forces identical rows. Floor attainment is equivalent to a unitary product realization of the row- and column-reduced table in dimension $n$ with orthonormal output matrices. In particular, after independent relabeling of its rows, columns, and outputs, the reduced table occurs as a subtable of a finite-group table. Targets lacking the required unitary realization have a strictly positive infimum gap above the universal floor.}

The resulting structural preference admits a { description-length interpretation. Group-isotope tables of order $n$ admit $\mathcal{O}(n\log n)$-bit descriptions, whereas almost all quasigroup tables require $\Theta(n^2\log n)$ bits. 
Beyond quasigroups, floor attainment supplies a finite-group realization of the reduced table, but its order is not uniformly controlled by the present results (Appendix~\ref{app:finite-group-realizations}).
Thus, the sensitivity objective favors
algebraic structures that support { economical descriptions},
without explicitly minimizing description length.
}

Numerical estimates of normalized algebraic complexity tend to increase with latent associativity violation $\nu(\delta)$. This association suggests a graded relationship between algebraic complexity and latent associativity beyond the exact floor-attainment characterization.

\paragraph{Broader ML Implications.}
Our results show how sensitivity minimization selects shared algebraic composition—providing a structural foundation for \emph{systematic generalization}: predicting novel combinations of known symbols by reusing the same compositional rule.
More broadly, these results reveal a simplicity bias toward \emph{functional compression} beyond low-rank interpolation, suggesting that sensitivity minimization could help deep networks acquire reusable compositional computation.

\paragraph{Limitations and Future Work.}
Our theoretical analysis concerns exact reconstruction of finite, fully observed operation tables. Extending the characterization to partial or noisy observations, deriving formal recovery rates, sample complexity bounds, and establishing convergence guarantees for gradient-based optimizers remain important open challenges.
A further question is how small a finite-group realization can be guaranteed for floor-attaining targets, and what uniform description-length bounds follow.
Future work will investigate comparable algebraic preferences in MLPs and Transformers under sensitivity minimization.
Beyond these settings, a broader direction is to characterize functional compression in high-dimensional, real-world continuous data and investigate its implications for generalization.

\newpage

\subsection*{AI use statement}

All research ideas, mathematical claims, and proofs in this paper were
developed by the authors without generative AI. We used generative AI tools
for two purposes. First, to help translate our proofs into the
Lean~4 formalization in \url{https://anonymous.4open.science/r/HyperCube_Lean-4664/README.md}. Second, to polish the
language, editing, and formatting of parts of the manuscript; these edits were reviewed
by the authors and introduced no technical content.

\bibliographystyle{iclr2027_conference}
\bibliography{references}

\newpage
\appendix

\section{Formal Definitions and Auxiliary Setup}

\subsection{Enumeration and Symmetry Classes}
\label{app:enumeration}

While the main text formulates the problem for general quasigroups, our experimental analysis (Appendix~\ref{app:experiments}) relies on the enumeration of loops to track the search space size.

\paragraph{Loops and Reduced Latin Squares.} 
A loop is a quasigroup with a two-sided identity element $e \in Q$. If we label the elements of $Q$ as $\{0, 1, \dots, n-1\}$ with $0=e$, the operation table is a \textit{reduced} (or normalized) Latin square, where the first row and first column are in natural order $(0, 1, \dots, n-1)$. Classifying loops of order $n$ is equivalent to classifying reduced Latin squares of order $n$.

\paragraph{Isomorphism vs. Isotopy.} 
The enumeration of distinct structures depends on the chosen notion of equivalence:
\begin{itemize}[leftmargin=2.5em, labelsep=0.5em]
    \item \textbf{Loop Isomorphism (Simultaneous Relabeling):} Two loops are isomorphic if they differ by a bijection $\phi: Q \to Q'$ satisfying $\phi(a) \circ' \phi(b) = \phi(a \circ b)$. In terms of Latin squares, this corresponds to applying the \textit{same} permutation to rows, columns, and symbol labels simultaneously. We use this relation to identify unique loops.
    
    \item \textbf{Quasigroup Isotopy (Independent Relabeling):} The model~\eqref{eq:hypercube}, which treats the operation as a 3-way tensor without a privileged identity, naturally respects a larger internal symmetry group. Two Latin squares are \textit{isotopic} if they differ by an arbitrary triple of permutations $(\phi, \psi, \chi) \in S_n \times S_n \times S_n$ acting independently on rows, columns, and symbols.
\end{itemize}
Since the sensitivity objective is exactly invariant under this larger group (as detailed in Section~\ref{sec:symmetries}), counting distinct structures for the optimization landscape corresponds to counting quasigroup isotopy classes (orbits of ${S_n}^3$).

\section{Deferred Proofs}
\label{app:deferred_proofs}

{
\begin{proposition}[Exact Factorization Exists]
\label{prop:feasible-nonempty}
Every binary operation target admits an exact factorization \eqref{eq:hypercube}: i.e., $\forall \delta$,  $\exists \Theta$ such that $T(\Theta)=\delta$.
\end{proposition}
This ensures that the feasible set  $\mathcal{F}_\delta $ (Definition~\ref{def:feasible_set}) is always non-empty. 

\begin{proof}[Proof of Proposition~\ref{prop:feasible-nonempty}]
Let $n=|Q|\geq1$, fix $q\in Q$, and let $E_{ij}$ denote the $n\times n$ matrix units indexed by $Q$. Set
\[
A_a=nE_{qa},\qquad
(B_b)_{ij}=\mathbb I_{\{i\circ b=j\}},\qquad
C_c=E_{cq}.
\]
Then $A_aB_bC_c=n(B_b)_{ac}E_{qq}$. Since $\Tr(E_{qq})=1/n$, normalized trace gives
\[
T_{abc}(\Theta)=\Tr(A_aB_bC_c)=(B_b)_{ac}=\delta_{abc}
\]
for every $a,b,c\in Q$, including unsupported triples. Thus, $\Theta\in\mathcal F_\delta$.
\end{proof}
}

\subsection{Proofs of Auxiliary Lemmas}

\begin{proof}[\textbf{Proof of Lemma~\ref{lem:decomposition}: Orthogonal Decomposition}]

Let the collinearity error between $A_a$ and $\partial T_{abc}/\partial A_a = (B_bC_c)^\dagger$ in \eqref{eq:alignment} be 
\begin{align*}
    \Delta_{abc}^{(A)} \coloneqq (B_b C_c)^\dagger - \frac{T_{abc}^*}{\|A_a\|^2} A_a            \end{align*}
Expanding its squared Frobenius norm yields:
\begin{align*}
    \|\Delta_{abc}^{(A)}\|^2 &= \left\langle (B_b C_c)^\dagger - \frac{T_{abc}^*}{\|A_a\|^2} A_a, \; (B_b C_c)^\dagger - \frac{T_{abc}^*}{\|A_a\|^2} A_a \right\rangle \\
    &= \|B_b C_c\|^2 - \frac{T_{abc}^*}{\|A_a\|^2} \langle (B_b C_c)^\dagger, A_a \rangle - \frac{T_{abc}}{\|A_a\|^2} \langle A_a, (B_b C_c)^\dagger \rangle + \frac{|T_{abc}|^2}{\|A_a\|^4} \|A_a\|^2 \\
    &= \|B_b C_c\|^2 - \frac{T_{abc} T_{abc}^*}{\|A_a\|^2} - \frac{T_{abc}^* T_{abc}}{\|A_a\|^2} + \frac{|T_{abc}|^2}{\|A_a\|^2} \\
    &= \|B_b C_c\|^2 - \frac{|T_{abc}|^2}{\|A_a\|^2}.
\end{align*}
Similarly, 
applying this expansion
 to $\Delta_{abc}^{(B)}$ and $\Delta_{abc}^{(C)}$ (misalignment for matrix $B$ and $C$), 
and summing over all supported triples yields
\begin{align*}
    \mathcal{R}_{\delta}(\Theta) 
    & \eqqcolon { 
    \sum_{\supp(\delta)} \left(
    \|\Delta_{abc}^{(A)}\|^2    +
    \|\Delta_{abc}^{(B)}\|^2    +
    \|\Delta_{abc}^{(C)}\|^2  \right) }  \\
    &=     
    \sum_{\supp(\delta)} \left( \|B_b C_c\|^2 + \|C_c A_a\|^2 + \|A_a B_b\|^2 - |T_{abc}|^2 \left( \frac{1}{\|A_a\|^2} + \frac{1}{\|B_b\|^2} + \frac{1}{\|C_c\|^2} \right) \right) \\
    &= \mathcal{H}(\Theta) - \mathcal{B}_{\delta}(\Theta).
\end{align*}
Rearranging this equation yields the exact decomposition $\mathcal{H}(\Theta) = \mathcal{B}_{\delta}(\Theta) + \mathcal{R}_{\delta}(\Theta)$. Since $\mathcal{R}_{\delta}$ is defined as a sum of squared norms, $\mathcal{R} \ge 0$, enforcing the lower bound $\mathcal{H} \ge \mathcal{B}_\delta $.
\end{proof}

\begin{proof}[\textbf{Proof of Lemma~\ref{lem:index-independent-gram}: Shared Gram Matrices}]
Substituting the collinear identities \eqref{eq:alignment} into 
the equation 
$B_b (C_c A_a) = (B_b C_c) A_a$ 
(i.e., associativity of matrix multiplication)
and canceling the nonzero scalar $T_{abc}$ 
isolates the relation 
$$Z = \frac{B_b B_b^\dagger}{\|B_b\|^2} = \frac{A_a^\dagger A_a}{\|A_a\|^2}.$$

{ which defines $Z$ locally for this supported triple.}
Note that the left side depends only on $b$, while the right side depends only on $a$.
Every input pair $(a,b)$ occurs { for any binary operation}, making $Z$ index-independent.
Taking the trace yields {$\Tr(Z) = \Tr(A_a A_a^\dagger)/\|A_a\|^2 = \|A_a\|^2/\|A_a\|^2 = 1$}.
The local Gram equalities for $X$ and $Y$ are obtained similarly.

{
More generally, form the bipartite support graphs on separate copies of $Q$ for the $(B,C)$ and $(A,C)$ roles. Their edges are $(b,c)$ whenever $\delta_{abc}=1$ for some $a$, and $(a,c)$ whenever $\delta_{abc}=1$ for some $b$, respectively. The local Gram equalities identify the normalized Gram matrices at adjacent vertices, so equality propagates along paths. Thus, $X$ and $Y$ are constant on each connected component containing an edge of the respective graphs.
}

{ For a quasigroup target,} $X$ and $Y$ are also shared across all supported triples, 
since every pair $(b,c)$ and every pair $(a,c)$
also occurs in a supported triple, yielding index-independent $X$ and $Y$, respectively,
{ but none of the subsequent lemmas requires their global sharing.}
\end{proof}

\begin{proof}[\textbf{Proof of Lemma~\ref{lem:proj-kappa}: Normalized Rank $\kappa$}]
Fix a supported triple $(a,b,c)$, and let $Z$ be the shared
normalized Gram matrix from Lemma~\ref{lem:index-independent-gram}.
Starting from \(Z = {A_a^\dagger A_a}/{\|A_a\|^2}\) and substituting the collinear relation $A_a^\dagger \propto B_b C_c$ from \eqref{eq:alignment} and the Gram identity for $X$ from \eqref{eq:gram-identities}:
{\small
\begin{align*}
  Z = \frac{\|A_a\|^2}{|T_{abc}|^2}\, (B_b C_c) (C_c^\dagger B_b^\dagger)
    = \frac{\|A_a\|^2\|C_c\|^2}{|T_{abc}|^2}\,  B_b \left( \frac{C_c C_c^\dagger}{\|C_c\|^2} \right) B_b^\dagger 
    = & {\frac{\|A_a\|^2 \|B_b\|^2 \|C_c\|^2}{|T_{abc}|^2}}\, \frac{(B_b B_b^\dagger) (B_b B_b^\dagger)}{\|B_b\|^2 \|B_b\|^2}
    \\
    = & {\kappa_{abc}} \, Z^2.
\end{align*}
}{
Since $Z$ is shared and $\Tr(Z) = 1 > 0$, we have $Z^2 \neq 0$.}
Thus, $Z = \kappa_{abc} Z^2$ determines the same coefficient
$\kappa_{abc}=\kappa$ for every { supported triple $(a,b,c)$.}
Defining $P \coloneqq \kappa Z$ yields $P^2 = \kappa^2 Z^2 = \kappa Z = P$, making $P$ an orthogonal projection, {forcing its eigenvalues into} a \textbf{{binary} set $\{0, 1\}$}. Thus, {$\operatorname{rank}(Z)
= \operatorname{rank}(P) = n\Tr(P) = n\kappa \Tr(Z) = \kappa \, n$},
which yields
$$  \kappa \, n = {\operatorname{rank}(Z)}
   = {\operatorname{rank}(A_a)} = {\operatorname{rank}(B_b)} 
    \le n .$$
Cyclically permuting this calculation gives the same projection identity for the local $X$ and $Y$,
yielding 
{ 
$$  \kappa \, n =  {\operatorname{rank}(C_c)} \le n $$
for $c$ that appear on the supported triple.
Unused output slices $C_c$ are unconstrained.}

At $\kappa=1$, each Gram matrix is a projection with normalized trace $1$ and hence $I_n$.
Conversely, $Z=I_n$ in $Z=\kappa Z^2$ gives $\kappa=1$.
\end{proof}

\begin{proof}[\textbf{Proof of Lemma~\ref{lem:AMGM}: AM-GM Lower Bound}]
Assume collinearity ($\mathcal{R}=0$) 
and $T_{abc} \neq 0$ on the support.
By Lemma~\ref{lem:decomposition}, $\mathcal H(\Theta)=\mathcal B_\delta(\Theta)$. All input and used output slices have nonzero norm, since each participates in a supported triple with nonzero prediction.
Let $\alpha_a \coloneqq \|A_a\|^{-2}$, $\beta_b \coloneqq \|B_b\|^{-2}$, and $\gamma_c \coloneqq \|C_c\|^{-2}$ { for participating slices}.
Recall from the definition of the normalized rank $\kappa$ (Lemma~\ref{lem:proj-kappa}) that $\kappa = \|A_a\|^2 \|B_b\|^2 \|C_c\|^2 / |T_{abc}|^2 \le 1$ on the support of $\delta$.
This implies the product identity: $\alpha_a \beta_b \gamma_c = \kappa^{-1} |T_{abc}|^{-2}$  for every supported triple.

Applying the scalar AM-GM inequality ($\alpha+\beta+\gamma \ge 3(\alpha\beta\gamma)^{1/3}$) to the definition of the base penalty $\mathcal{B}_{\delta}$ yields:
\begin{align*}
    \mathcal H(\Theta)=\mathcal{B}_{\delta}(\Theta) &= \sum_{\supp(\delta)} |T_{abc}|^2(\alpha_a+\beta_b+\gamma_c) \\
    &\ge \sum_{\supp(\delta)} |T_{abc}|^2 \cdot 3(\alpha_a\beta_b\gamma_c)^{1/3} \\
    &= 3 \kappa^{-1/3} \sum_{\supp(\delta)} \, |T_{abc}|^{4/3} \\
    & \ge \mathcal H^*_\delta(T(\Theta)).
\end{align*}

Equality in the first inequality holds \emph{if and only if} the AM-GM condition $\alpha_a = \beta_b = \gamma_c$ is satisfied for every supported triple $(a,b,c)$. Indeed, every supported weight $|T_{abc}|^2$ is positive, so equality in the sum requires equality in every scalar AM-GM inequality. Since $(a,b,a\circ b)$ is supported for every input pair $(a,b)$, these equalities make all input norms equal; every used output norm then has the same value. Conversely, this balance condition gives equality in each scalar inequality. { The equality condition imposes no restriction on unused output norms.}

{ Finally, $0<\kappa\le1$ gives $\kappa^{-1/3}\ge1$, proving the second inequality. Since $\mathcal H^*_\delta(T(\Theta))>0$, equality in this step holds if and only if $\kappa=1$. Thus $\mathcal H(\Theta)=\mathcal H^*_\delta(T(\Theta))$ holds if and only if the participating norms are balanced and $\kappa=1$.}
\end{proof}

\subsection{Proof of Theorem~\ref{thm:unitary_equivalence}: Unitary Collinearity--Associativity Equivalence}
\label{app:unitary_proof}

To formally prove Theorem~\ref{thm:unitary_equivalence}, we first establish supporting lemmas regarding gauge synchronization, the resulting homomorphism, and the constructive sufficiency of the left-regular representation. By isotopy equivariance (Section~\ref{sec:symmetries}), we assume without loss of generality that the target $\delta$ is a loop $(Q, \circ')$ (i.e. a quasigroup with an identity $e$).

\begin{lemma}[Synchronization]
\label{lem:app_synchronization}
Let $(Q, \circ')$ be a finite loop admitting a unitary collinear factorization $\Theta$. There exists a unitary gauge transformation $(U, V, W)$ such that the transformed parameter triple $\Theta' = (A',B',C')$ {{ remains feasible} and satisfies} the synchronized condition: $A'_g = B'_g = (C'_g)^\dagger \coloneqq \rho(g)$ for all $g \in Q$.
\end{lemma}
\begin{proof}
Since the factorization is unitary, the slices $A_e, B_e$ are unitary matrices. We explicitly define 
the unitary gauge as $(U, V, W) \coloneqq (A_e^\dagger, I_n, B_e)$.
{ Feasibility gives $T_{abc}=1$ on supported triples; together with collinearity and unitarity, \eqref{eq:alignment} therefore gives the exact identity} 
$A_a B_b = C_c^\dagger$.
Applying this to triples $(g, e, g)$ and $(e, g, g)$ yields $A_g B_e = C_g^\dagger$ and $A_e B_g = C_g^\dagger$, implying $A_g B_e = A_e B_g$.
Applying the gauge transformation:
\begin{align*}
    A'_g &= A_e^\dagger A_g = A_e^\dagger (A_e B_g B_e^\dagger) = B_g B_e^\dagger = B'_g.
\end{align*}
For the third factor:
\begin{align*}
    C'_g &= B_e C_g A_e = B_e (B_e^\dagger A_g^\dagger) A_e = A_g^\dagger A_e = (A'_g)^\dagger.
\end{align*}
Thus, all three factors collapse to a single unitary map $\rho(g)$.
{ Gauge invariance preserves $T(\Theta')=T(\Theta)=\delta$ and the objective. Writing $P=A_e$ and $Q=B_e$, the original factors are recovered exactly as
\[
A_g=P\rho(g),\qquad B_g=\rho(g)Q,\qquad
C_g=Q^\dagger\rho(g)^\dagger P^\dagger.
\]}
\end{proof}

\begin{lemma}[Homomorphism and Injectivity]
\label{lem:app_homomorphism}
Let $\rho(g)$ be the synchronized map from Lemma~\ref{lem:app_synchronization}. The collinearity condition enforces that $\rho$ is a homomorphism. Furthermore, the feasibility condition $T(\Theta)=\delta$ ensures $\rho$ is an injective map.
\end{lemma}
\begin{proof}
By the collinear identities of the transformed factors, $A'_a B'_b = T'_{abc} (C'_c)^\dagger$ for supported triples ($a \circ' b = c$).
Substituting the synchronized map yields $\rho(a)\rho(b) = T'_{abc} \rho(c)$. 
Because the factorization is unitary and feasible, the trace inner product evaluates to $T'_{abc} = 1$. Thus, $\rho(a)\rho(b) = \rho(c) = \rho(a \circ' b)$, establishing $\rho$ as a true group homomorphism into $U(n)$.

To prove injectivity, suppose $\rho(x) = \rho(y)$ for some $x, y \in Q$. For the valid triple $(x, e, x)$, feasibility requires $T'_{xex} = \Tr(\rho(x)\rho(e)\rho(x)^\dagger) = 1$.
Because $\rho(x)=\rho(y)$, substituting $y$ yields:
$T'_{yex} = \Tr(\rho(y)\rho(e)\rho(x)^\dagger) = \Tr(\rho(x)\rho(e)\rho(x)^\dagger) = 1$.
Feasibility ($T'_{yex} = \delta_{yex}$) dictates that $\delta_{yex} = 1$, which means $y \circ' e = x$. Since $e$ is the identity, $y \circ' e = y$, forcing $x = y$. Thus, $\rho$ is injective.
\end{proof}

\begin{proof}[\textbf{Proof of Theorem~\ref{thm:unitary_equivalence}: Unitary Collinearity $\iff$ Group Isotope}]
(\textbf{Necessity}): 
As established via isotopy equivariance, 
if $(Q, \circ)$ admits a unitary collinear factorization, so does its loop isotope $(Q, \circ')$. 
By Lemma~\ref{lem:app_synchronization}, this factorization induces a synchronized map $\rho: Q \to U(n)$.
Since $\rho$ is a homomorphism (Lemma~\ref{lem:app_homomorphism}), invoking the associativity of continuous matrix multiplication yields:
\begin{equation*} 
    \rho((x \circ' y) \circ' z) = \rho(x)\rho(y)\rho(z) = \rho(x \circ' (y \circ' z)). 
\end{equation*}
Since $\rho$ is injective (Lemma~\ref{lem:app_homomorphism}), this equality implies the associativity of the discrete operation $\circ'$:  $(x \circ' y) \circ' z = x \circ' (y \circ' z)$. 
Thus, the loop $(Q, \circ')$ is an associative group, and the original quasigroup $(Q, \circ)$ is a group isotope.

\noindent(\textbf{Sufficiency}): 
This direction is established constructively in Lemma~\ref{lem:app_group_existence}. \end{proof}

\begin{lemma}[Uniqueness of Representation]
\label{lem:app_representation_uniqueness}
Let $(Q,\circ')$ be the group identified in the necessity proof. The induced map $\rho: Q \to U(n)$ is unitarily equivalent to the left-regular representation of $(Q,\circ')$.
\end{lemma}
\begin{proof}
Since $\rho$ is a true group homomorphism (Lemma~\ref{lem:app_homomorphism}), we can evaluate the feasibility constraint on the specific triple $(g, e, e)$:
\begin{equation*}
    T_{gee} = \Tr(\rho(g)\rho(e)\rho(e)^\dagger) = \Tr(\rho(g \circ' e \circ' e^{-1})) = \Tr(\rho(g)) = \delta_{gee}.
\end{equation*}
Thus, the character of $\rho$ evaluates to: $\operatorname{Trace}(\rho(g)) = n \cdot \delta_{gee} = n \cdot \mathbb{I}_{\{g=e\}}$,
which is precisely the character of the left-regular representation 
$\lambda$. 
{ Standard character theory first supplies an invertible intertwiner $S$ with $S\lambda(g)=\rho(g)S$. Since both representations are unitary, $S^\dagger S$ commutes with every $\lambda(g)$; polar normalization $K=S(S^\dagger S)^{-1/2}$ gives a unitary intertwiner. Hence $\rho(g)=K\lambda(g)K^\dagger$ for all $g$.

Combining this with Lemma~\ref{lem:app_synchronization}, set $L=PK$, $R=K$, and $W=Q^\dagger K$. These matrices are unitary, and
\[
A_g=L\lambda(g)R^\dagger,\qquad
B_g=R\lambda(g)W^\dagger,\qquad
C_g=W\lambda(g)^\dagger L^\dagger.
\]
Undoing the loop isotopy yields independent permutations $\phi,\psi,\chi$ satisfying $\phi(a)\circ'\psi(b)=\chi(a\circ b)$ and the same formulas with $\lambda(\phi(a))$, $\lambda(\psi(b))$, and $\lambda(\chi(c))$ in the three respective families. Thus the classification reconstructs the original factorization, not only its synchronized representation. For a group-isotope target, Theorem~\ref{thm:absolute_lower_bound} and Lemma~\ref{lem:app_group_existence} imply that every global minimizer attains the floor and is unitary collinear, so this classification applies to every such minimizer.
}
\end{proof}

\begin{lemma}[Sufficiency: Group Isotope $\implies$ Unitary Collinear Factorization]
\label{lem:app_group_existence}
Let $(Q, \circ)$ be a quasigroup isotopic to a group $(Q, \circ')$. Then, $(Q, \circ)$ admits a unitary collinear factorization ($\mathcal{F}_\delta \cap \mathcal{M}_\delta \cap \mathcal{U}_\delta$) and attains an objective value of 
$\mathcal{H}(\Theta) = 3 \, |\supp(\delta)|$.
\end{lemma}

\begin{proof}

Let $\rho: Q \to U(n)$ be the left-regular representation of the group $(Q, \circ')$. By definition, $\rho(g)$ is a unitary matrix for all $g \in Q$. 
Define a synchronized unitary parameter triple $\Theta \in \mathcal{U}_\delta$ as:
$A_g = B_g = C_g^\dagger = \rho(g)$.

For any supported triple $a \circ' b = c$, we have $A_a B_b = \rho(a)\rho(b) = \rho(a \circ' b) = \rho(c) = C_c^\dagger$,
satisfying collinearity ($\mathcal{M}_\delta$) \eqref{eq:alignment}. 
Also, their product \eqref{eq:hypercube} satisfies feasibility ($\mathcal{F}_\delta$).
\begin{equation*}
    T_{abc} (\Theta) = \Tr(\rho(a)\rho(b)\rho(c)^\dagger) = \Tr(\rho(c)\rho(c)^\dagger) = \Tr(I_n) = 1 = \delta_{abc}.
\end{equation*}
{ More generally, $\Tr(\rho(g))=\mathbb{I}_{\{g=e\}}$ gives $T_{abc}=\mathbb{I}_{\{a\circ' b=c\}}$ for all triples, and isotopy equivariance transfers this factorization to the original target.}
For any unitary parameters $\Theta \in \mathcal{U}_\delta$, 
the objective \eqref{eq:H-Latin} evaluates to 
\begin{equation*}
    \mathcal{H}(\Theta) = \sum_{\supp(\delta)} (\|B_b C_c\|^2 + \|C_c A_a\|^2 + \|A_a B_b\|^2) = \sum_{\supp(\delta)} (1^2 + 1^2 + 1^2) = 3 \, |\supp(\delta)|.
\end{equation*}
\end{proof}

\section{Matrix AM-GM Inequality { and Uniform Positive Infimum Gap}}
\label{app:matrix-AM-GM}

Throughout this section, we extend the normalized trace and the normalized Frobenius norm from the main text to square matrices of arbitrary dimension $N$: for $M \in \mathbb{C}^{N \times N}$, we write $\Tr(M)\coloneqq \frac{1}{N}\text{Trace}(M)$ and $\|M\|^2 \coloneqq \Tr(M^\dagger M)$.

\subsection{Supporting Lemma}

\begin{lemma}[Spectral Trace--Frobenius Bound]
\label{lem:spectral_trace_frob}
For every $M\in\mathbb{C}^{N\times N}$, the normalized trace and Frobenius norm satisfy
\begin{equation}
    \|M^2\|^2  = 
    \Tr({M^2}^\dagger M^2)
    \ge 
    \left| \Tr(M^3) \right|^{4/3} .
        \label{eq:spectral_trace_frob}
\end{equation}
\end{lemma}
\begin{proof}
Let $\lambda_1,\ldots,\lambda_N$ be the eigenvalues of $M$, counted with algebraic multiplicity, and set $r_i\coloneqq|\lambda_i|$. Since the eigenvalues of $M^3$ are $\lambda_i^3$, we have $\Tr(M^3)=\frac 1 N \sum_{i=1}^N \lambda_i^3$. Hence,
\begin{align}
    && \left| \Tr(M^3) \right|^{4/3}
    &= \left|\frac 1 N\sum_{i=1}^N \lambda_i^3 \right|^{4/3} \nonumber \\
    &&&\le \left(\frac1N\sum_{i=1}^N r_i^3 \right)^{4/3} \nonumber \\
    &&&\le \frac1N\sum_{i=1}^N r_i^4 &&\text{(Jensen's inequality)} 
        \label{eq:Jensen}
    \\
    &&&= \frac1N\sum_{i=1}^N |\lambda_i^2|^2 \nonumber \\
    &&&\le \|M^2\|^2. &&\text{(Schur/Frobenius domination)} 
    \label{eq:Schur_Frobenius}
\end{align}
The last inequality follows by applying Schur triangularization to $M^2$. If $Q^\dagger M^2Q=T$ is upper triangular with diagonal entries $\lambda_i^2$, then
\[
\|M^2\|^2 = \Tr(T^\dagger T) = \frac{1}{N}\left(\sum_i |\lambda_i^2|^2+\sum_{i<j}|T_{ij}|^2\right) \ge \frac{1}{N}\sum_i |\lambda_i^2|^2.
\]
Equality in this Schur/Frobenius step holds if and only if the Schur form has no strictly upper triangular part, equivalently if and only if $M^2$ is normal.
\end{proof}

\subsection{A Noncommutative Trace--Frobenius Matrix AM-GM Inequality}

The following matrix inequality replaces the scalar AM-GM argument. Crucially, it applies directly to arbitrary matrices without assuming commutativity.

\textbf{Matrix AM-GM Lemma~\ref{lem:matrix_AMGM}}
(Restated):  
For all $X,Y,Z \in \mathbb{C}^{n \times n}$,
\begin{equation}
    \|XY\|^2 + \|YZ\|^2 + \|ZX\|^2 \ge 3 \, |\Tr(XYZ)|^{4/3}.
    \end{equation}
If $t\coloneqq\Tr(XYZ)\neq0$, equality holds if and only if
\[
\frac{XX^\dagger}{\|X\|^2}
=\frac{YY^\dagger}{\|Y\|^2}
=\frac{ZZ^\dagger}{\|Z\|^2}=I_n,
\qquad
\|X\|^2=\|Y\|^2=\|Z\|^2,
\qquad XYZ=tI_n.
\]
These conditions imply
$\kappa\coloneqq\|X\|^2\|Y\|^2\|Z\|^2/|t|^2=1$.
{ If $t=0$, equality holds if and only if $XY=YZ=ZX=0$.}

\begin{proof}
Define the block-cyclic matrix
\[
M \coloneqq
\begin{pmatrix}
0 & X & 0 \\
0 & 0 & Y \\
Z & 0 & 0
\end{pmatrix}\in\mathbb{C}^{3n\times 3n}.
\]
Then
\[
M^2=
\begin{pmatrix}
0 & 0 & XY \\
YZ & 0 & 0 \\
0 & ZX & 0
\end{pmatrix},
\qquad
M^3=
\begin{pmatrix}
XYZ & 0 & 0 \\
0 & YZX & 0 \\
0 & 0 & ZXY
\end{pmatrix}.
\]
Evaluating the normalized norm of $M^2$ yields:
\[
\|M^2\|^2 = \frac{1}{3n}\text{Trace}\left((M^2)^\dagger M^2\right) = \frac13\bigl(\|XY\|^2+\|YZ\|^2+\|ZX\|^2\bigr).
\]
Similarly, by the cyclicity of the standard trace, $\Tr(YZX) = \Tr(ZXY) = \Tr(XYZ)$. Evaluating the normalized trace of $M^3$ yields:
\[
\Tr(M^3) = \frac{1}{3n}\operatorname{Trace}(M^3) = \frac{1}{3n}\bigl(3 \operatorname{Trace}(XYZ)\bigr) = \Tr(XYZ).
\]
Substituting $M$ into the spectral bound from Lemma~\ref{lem:spectral_trace_frob} gives exactly \eqref{eq:matrix_amgm}.

{To prove the equality condition, we first prove the normalized case $\Tr(XYZ)=1$; we then recover arbitrary nonzero trace by rescaling and treat zero trace separately.}

{Suppose $\Tr(XYZ)=1$ and equality holds in \eqref{eq:matrix_amgm}. Then $\Tr(M^3)=1$ and}
\[
\|M^2\|^2 = \frac13\bigl(\|XY\|^2+\|YZ\|^2+\|ZX\|^2\bigr)=1.
\]
Thus equality holds at every step of the spectral chain above for $M$.
Let $\lambda_1, \ldots,\lambda_{3n}$ be the eigenvalues of $M$ and put $r_i=|\lambda_i|$.

First, equality in the Jensen's step \eqref{eq:Jensen}
\[
\left(\frac1{3n}\sum_i r_i^3\right)^{1/3} \le \left(\frac1{3n}\sum_i r_i^4\right)^{1/4}
\]
forces all $r_i$ to be equal, say $r_i=\rho$ for every $i$. Since equality also holds in the triangle step and $\Tr(M^3)=1$, we have
\[
1=\left|\frac1{3n}\sum_i \lambda_i^3\right| = \frac1{3n}\sum_i |\lambda_i|^3 = \rho^3.
\]
Hence $\rho=1$; all eigenvalues of $M$ lie on the unit circle.

Second, equality in the triangle inequality implies that the complex numbers $\lambda_i^3$ all have the same argument. Since $|\lambda_i^3|=1$ and their average is the real number $1$,
\[
\frac1{3n}\sum_i \lambda_i^3=1,
\]
the common argument must be zero. Therefore
\begin{equation}
    \lambda_i^3=1 \qquad \text{for every } i.
    \label{eq:eigs_cube_roots}
\end{equation}

Third, equality in the Schur/Frobenius domination step \eqref{eq:Schur_Frobenius} forces $M^2$ to be normal. The eigenvalues of $M^2$ are $\lambda_i^2$, and by the preceding paragraph they all have modulus one. A normal matrix with all eigenvalues on the unit circle is unitary; hence
\begin{equation}
    M^2 \text{ is unitary.}
    \label{eq:M2_unitary}
\end{equation}
Indeed, after unitary diagonalization of $M^2$, the relation $(M^2)^\dagger M^2=I$ follows entrywise from $|\lambda_i^2|=1$.

We next show that $M^3=I_{3n}$, not merely that the eigenvalues of $M^3$ are equal to $1$. Since $M^2$ is normal and its eigenvalues $\lambda_i^2$ satisfy $(\lambda_i^2)^3=1$, functional calculus for normal matrices gives
\[
(M^2)^3=I_{3n},
\]
so $M^6=I_{3n}$. The polynomial $t^6-1$ has no repeated roots, so any matrix annihilated by it is diagonalizable. Thus $M$ is diagonalizable. Combining diagonalizability with \eqref{eq:eigs_cube_roots} gives
\begin{equation}
    M^3=I_{3n}.
    \label{eq:M3_identity}
\end{equation}

Now inspect the block consequences of \eqref{eq:M2_unitary} and \eqref{eq:M3_identity}. From the displayed form of $M^2$, unitarity of $M^2$ implies that each block product $XY$, $YZ$, and $ZX$ is unitary. From the displayed form of $M^3$, the identity \eqref{eq:M3_identity} gives
\begin{equation}
    XYZ=I, \qquad YZX=I, \qquad ZXY=I.
    \label{eq:cyclic_identities}
\end{equation}
Finally, these facts force the individual factors to be unitary. Since $XY$ is unitary and $XYZ=I$, we have
\[
Z=(XY)^{-1}=(XY)^\dagger,
\]
so $Z$ is unitary. Since $YZ$ is unitary and $Z$ is unitary,
\[
Y=(YZ)Z^\dagger
\]
is unitary. Since $XY$ is unitary and $Y$ is unitary,
\[
X=(XY)Y^\dagger
\]
is unitary. Hence equality implies $X,Y,Z$ are unitary and $XYZ=I$.

Conversely, if $X,Y,Z$ are unitary and $XYZ=I$, then $XY$, $YZ$, and $ZX$ are unitary, so each has normalized Frobenius norm squared equal to $1$, while $|\Tr(XYZ)|^{4/3}=|\Tr(I)|^{4/3}=1$. Thus both sides of \eqref{eq:matrix_amgm} equal $3$, proving equality.

{
Now let $t=\Tr(XYZ)\neq0$, and choose $\tau\in\mathbb C$ with $\tau^3=t$. Set $\widehat X=X/\tau$, $\widehat Y=Y/\tau$, and $\widehat Z=Z/\tau$. Then
\[
\Tr(\widehat X\widehat Y\widehat Z)=1,\qquad
\|\widehat X\widehat Y\|^2+\|\widehat Y\widehat Z\|^2+\|\widehat Z\widehat X\|^2
=\frac{\|XY\|^2+\|YZ\|^2+\|ZX\|^2}{|\tau|^4}.
\]
Since $|t|^{4/3}=|\tau|^4$, equality for the original triple is equivalent to equality at unit trace for the rescaled triple. The case just proved gives unitary $\widehat X,\widehat Y,\widehat Z$ and $\widehat X\widehat Y\widehat Z=I_n$. Rescaling back yields
\[
XX^\dagger=YY^\dagger=ZZ^\dagger=|\tau|^2I_n=|t|^{2/3}I_n,
\qquad XYZ=\tau^3I_n=tI_n.
\]
Conversely, these identities make the rescaled factors unitary with product identity, so equality follows. For positive real trace $t$, choosing $\tau=t^{1/3}>0$ recovers the stated positive-scale form.

Taking normalized traces gives
$\|X\|^2=\|Y\|^2=\|Z\|^2=|t|^{2/3}$.
Thus the normalized Gram matrices are $I_n$, the factors
are balanced, and
$\kappa=\|X\|^2\|Y\|^2\|Z\|^2/|t|^2=1$.
Together with $XYZ=tI_n$, these conditions imply the
collinear identities, recovering the full-rank, balanced
equality case of Lemma~\ref{lem:AMGM}.

Finally, if $t=0$, the right side of \eqref{eq:matrix_amgm} is zero. Its left side is a sum of nonnegative squared norms, so equality is equivalent to $XY=YZ=ZX=0$. This branch can include nonzero factors and need not consist of scaled unitaries.}
\end{proof}

{
\subsection{No-Escape and the Positive Infimum Gap}
\label{app:exact-fit-gap}

This section establishes Lemma~\ref{lem:spectral-no-escape} and Theorem~\ref{thm:positive-infimum-gap}.
We retain the normalized Frobenius norm: for $M\in\mathbb C^{d\times d}$, $\Tr(M) \coloneqq \frac 1 d \operatorname{Trace}(M)$ and $\|M\|^2 \coloneqq \Tr(M^\dagger M)$, so $\|I_d\|^2=1$. The normalizing dimension is $n$ for factor slices and $3n$ for block-cyclic matrices.
No global coercivity or existence of a nongroup minimizer is assumed.

\subsubsection{Spectral No-Escape Lemma}

\begin{lemma}[Spectral No-Escape]
\label{lem:spectral-no-escape}
Fix $d\geq1$, and let $M_k\in\mathbb C^{d\times d}$ satisfy
\[
\Tr(M_k^3) = 1,\qquad \|M_k^2\|^2\longrightarrow1.
\]
Then, $\sup_k\|M_k\|<\infty$, where
$\|M\|^2= \Tr(M^\dagger M)$.
\end{lemma}

\begin{proof}[Proof of Lemma~\ref{lem:spectral-no-escape}]
We first record the scalar equality observation needed for the Schur diagonal. If $z_1, \ldots ,z_d\in\mathbb C$ satisfy $\sum_i z_i^3=d$ and $\sum_i|z_i|^4 \leq  d$, then
\[
d=\left|\sum_i z_i^3\right|
 \leq \sum_i|z_i|^3
 \leq  d^{1/4}\left(\sum_i|z_i|^4\right)^{3/4}
 \leq  d.
\]
Thus, all inequalities are equalities. Equality in the middle inequality (H\"older, equivalently the Jensen step in Lemma~\ref{lem:spectral_trace_frob}) forces equal moduli, and equality in the last bound makes each modulus $1$. Since $\sum_i\operatorname{Re}(z_i^3)=d$ and each real part is at most $1$, every $z_i^3$ has real part and modulus $1$. Thus $z_i^3=1$ for every $i$.

Suppose now that a sequence satisfying the lemma were unbounded. Choose a subsequence with  $\|M_k\|^2\to\infty$ and relabel it; this divergence is preserved by subsequent subsequences. Schur triangularization gives unitary matrices $U_k$ and upper triangular matrices $T_k=U_k^\dagger M_k U_k$. Hence
\[
\|T_k\|^2\to\infty,\qquad \Tr(T_k^3)=1, \qquad \|T_k^2\|^2 \to 1.
\]
Choose $L \geq 1$ such that $d\|T_k^2\|^2\leq L$ for all $k$. For $\lambda_{k,i}=(T_k)_{ii}$, triangularity gives
\[
\sum_i\lambda_{k,i}^3=d,\qquad
\sum_i|\lambda_{k,i}|^4 \leq d\|T_k^2\|^2 \leq L.
\]
The diagonal vectors are bounded, so a further subsequence converges to $(\zeta_1,\ldots,\zeta_d)$. Passing to the limit gives $\sum_i\zeta_i^3=d$ and $\sum_i|\zeta_i|^4 \leq  d$. The scalar observation implies $\zeta_i^3=1$ for every $i$.

No two cube roots of unity sum to zero: if $\zeta_j=-\zeta_i$, their cubes would be negatives of one another. Set
\[
c=\tfrac12\min_{i,j}|\zeta_i+\zeta_j|>0.
\]
Convergence and finiteness give a common $K$ such that
\begin{equation}
\label{eq:schur-diagonal-sum-bound}
|\lambda_{k,i}+\lambda_{k,j}| \geq c\qquad(k \geq K,\ 1 \leq  i,j \leq  d).
\end{equation}

Set $D=L^{1/4}$ and $P=L^{1/2}$. The preceding bounds give
$|(T_k)_{ii}|\leq D$ and $|(T_k^2)_{ij}|\leq P$.
For $i<j$, upper triangularity yields
\begin{equation}
\label{eq:schur-superdiagonal-recurrence}
(T_k^2)_{ij}=(\lambda_{k,i}+\lambda_{k,j})(T_k)_{ij}
+\sum_{i<\ell<j}(T_k)_{i\ell}(T_k)_{\ell j}.
\end{equation}

We now prove a uniform entrywise bound by induction on $s$, the largest
separation $j-i$ controlled at the current stage. Set $B_0=D$ and
$B_{s+1}=\max\{B_s,(P+dB_s^2)/c\}$. We claim that, for each
$s=0,\ldots,d-1$,
\[
|(T_k)_{ij}|\leq B_s
\qquad(0\leq j-i\leq s,\ k\geq K).
\]
The case $s=0$ is the diagonal bound $|(T_k)_{ii}|\leq D=B_0$.

Suppose the claim holds for some $s<d-1$, and let $j-i=s+1$.
For $i<\ell<j$, both separations $\ell-i$ and $j-\ell$ are at most $s$,
so the induction hypothesis bounds each factor in the sum in
\eqref{eq:schur-superdiagonal-recurrence} by $B_s$.
Rearranging that identity and applying the triangle inequality gives
\[
\begin{aligned}
|\lambda_{k,i}+\lambda_{k,j}|\,|(T_k)_{ij}|
&\leq |(T_k^2)_{ij}|
 +\sum_{i<\ell<j}|(T_k)_{i\ell}|\,|(T_k)_{\ell j}|\\
&\leq P+dB_s^2,
\end{aligned}
\]
since the sum has at most $d$ terms. The uniform lower bound
\eqref{eq:schur-diagonal-sum-bound} therefore gives
\[
|(T_k)_{ij}|\leq (P+dB_s^2)/c\leq B_{s+1}.
\]
Since $B_{s+1}\geq B_s$, the same bound holds for entries of smaller
separation, completing the induction on the same tail $k\geq K$.

Taking $s=d-1$ bounds every upper triangular entry by $B_{d-1}$.
The remaining entries vanish, so
\[
\|T_k\|^2=d^{-1}\sum_{i,j}|(T_k)_{ij}|^2
\leq dB_{d-1}^2\qquad(k\geq K),
\]
contradicting the assumed divergence of $\|T_k\|$.

\end{proof}

\paragraph{Application to matrix triples.}
Suppose $X_k,Y_k,Z_k\in\mathbb C^{n \times n}$ satisfy
\[
\Tr(X_kY_kZ_k)=1,\qquad
 \|X_k Y_k\|^2+\|Y_k Z_k\|^2+\|Z_k X_k\|^2 \longrightarrow 3.
\]
Let $M_k$ be the block-cyclic matrix from the proof of Lemma~\ref{lem:matrix_AMGM}, with blocks $X_k,Y_k,Z_k$. Direct multiplication and trace cyclicity give $\Tr(M_k^3)=\Tr(X_k Y_k Z_k)=1$,
\[
\|M_k^2\|^2=\tfrac13\bigl(\|X_kY_k\|^2+\|Y_kZ_k\|^2+\|Z_kX_k\|^2\bigr) \longrightarrow 1,
\qquad
\|M_k\|^2=\tfrac13\bigl(\|X_k\|^2+\|Y_k\|^2+\|Z_k\|^2\bigr).
\]
Lemma~\ref{lem:spectral-no-escape} in dimension $3n$ therefore bounds all three factors. This is the local no-escape implication used below.

\subsubsection{Positive Infimum Gap}
\label{app:positive_gap}

\begin{proof}[Proof Sketch of Theorem~\ref{thm:positive-infimum-gap}]
If feasible objective values approach $3 \, |\supp(\delta)|$, every supported local sensitivity approaches $3$. Lemma~\ref{lem:spectral-no-escape}, applied to the corresponding block-cyclic matrices, bounds all input and used output slices; zeroing unused output slices preserves both feasibility and the objective, yielding a bounded parameter sequence. A convergent subsequence then has a feasible limit at the floor by closedness and continuity. Consequently, every target that does not attain the floor has a strictly positive target-specific gap. Taking the minimum of these gaps over the finitely many non-attaining binary-operation tables of order $n$ gives $\varepsilon_n$.
\end{proof}

\begin{proof}[Proof of Theorem~\ref{thm:positive-infimum-gap}]
Fix any binary operation $(Q,\circ)$ of order $n$, with structure tensor $\delta$, and choose $o\in Q$. The parameter space is a finite-dimensional real vector space of dimension $6 n^3$. Full feasibility defines a closed subset $\mathcal F_\delta$, and $\mathcal H$ is continuous, since both are defined by polynomials in the real and imaginary parts of matrix entries. Proposition~\ref{prop:feasible-nonempty} ensures that $\mathcal F_\delta$ is nonempty.

Let $U_\circ=\{a\circ b:a,b\in Q\}$. For each feasible $\Theta=(A,B,C)$, define $\widehat\Theta=(A,B,\widehat C)$ by
\[
\widehat C_c=\begin{cases}C_c,&c\in U_\circ,\\0,&c\notin U_\circ.\end{cases}
\]
This preserves full feasibility: used-output predictions are unchanged, while predictions for unused outputs are zero both before and after normalization. It also preserves $\mathcal H$, since every term in the support-weighted objective uses an output $a\circ b\in U_\circ$.

First consider any sequence $\Theta_k\in\mathcal F_\delta$ with $\mathcal H(\Theta_k) \to 3 \, |\supp(\delta)|$. For a supported triple define
\[
r_k(a,b)=\|A_a^{(k)}B_b^{(k)}\|^2
+\|B_b^{(k)}C_{a \circ b}^{(k)}\|^2
+\|C_{a \circ b}^{(k)}A_a^{(k)}\|^2.
\]
Matrix AM--GM and feasibility give $r_k(a,b)\geq3$ for every pair.
Since $\mathcal H(\Theta_k)=\sum_{u,v\in Q}r_k(u,v)$, each local
excess is bounded by the total:
\[
0\leq r_k(a,b)-3
\leq \mathcal H(\Theta_k)-3 \, |\supp(\delta)|\longrightarrow0.
\]
The local no-escape implication bounds each supported triple's factors. There are only finitely many input pairs. The pairs $(a,o)$ cover every $A_a$, the pairs $(o,b)$ cover every $B_b$, and every used output $c\in U_\circ$ has an observed pair $(a,b)$ with $a\circ b=c$. The remaining $\widehat C_c$ slices are zero. Taking a maximum over these finitely many bounds proves boundedness of the normalized sequence $\widehat\Theta_k$ in parameter space. 

Suppose $\delta$ admits no feasible factorization at the floor and no positive target-specific gap exists. The universal bound then permits a sequence in $\mathcal F_\delta$ with
\[
3 \, |\supp(\delta)| \leq \mathcal H(\Theta_k) < 3 \, |\supp(\delta)| + \frac 1 {k+1}.
\]
Since the parameter space is finite-dimensional, the bounded sequence $\widehat\Theta_k$ has a subsequence converging to some $\Theta_*$.
Normalization preserves feasibility and the objective, so closedness and continuity imply $\Theta_*\in\mathcal F_\delta$ and $\mathcal H(\Theta_*)=3 \, |\supp(\delta)|$, contradicting the assumed nonattainment of the floor. Therefore, some $\varepsilon_\circ>0$ satisfies $\mathcal H(\Theta) \geq 3 \, |\supp(\delta)|+\varepsilon_\circ$ for every feasible $\Theta$.

Finally, at fixed $n$ there are only $n^{n^2}$ binary operations on a labeled $n$-element set, hence finitely many targets that do not attain the floor. If this eligible set is nonempty, choose the minimum of their positive constants $\varepsilon_\circ$ as $\varepsilon_n$. If it is empty, take $\varepsilon_n=1$ and every target satisfies the floor-attainment alternative. This constant bounds every eligible feasible value; taking infima preserves the same bound.
\end{proof}

}

\section{Empirical Verification}
\label{app:experiments}

\subsection{Training Setting and Hyperparameters}
\label{app:experiments_hyperparameters}

To ensure reproducibility and provide a granular view of the learning dynamics discussed in the main text, we detail the specific training hyperparameters used to generate the optimization trajectories. We optimize the model by minimizing the squared-error reconstruction loss augmented by the sensitivity objective:
\begin{equation}
    \mathcal{L}(\Theta) = \|T(\Theta) - \delta\|^2_F + \lambda \mathcal{H}(\Theta)
\end{equation}
We track the objective components, \emph{i.e.}, 
$\mathcal{H}, \mathcal{B}_{\delta}, \mathcal{R}_{\delta}, \mathcal{H}_{\delta}^*$,
and the $\ell_2$ norm  ($\|\Theta\|^2$) for reference. 

We use the standard Gradient Descent optimizer (learning rate $0.5$, zero momentum) with a two-phase schedule for $\lambda$. 
We set $\lambda=0.05$ for the first $500$ epochs,
then set $\lambda=0$ to fit the target exactly.
The normalized objective components $\tilde{\mathcal B}_\delta\coloneqq\mathcal B_\delta/\mathcal H^*_\delta$ and $\tilde{\mathcal R}_\delta\coloneqq\mathcal R_\delta/\mathcal H^*_\delta$ are compared before and after this transition.

\subsection{Detailed Optimization Dynamics 
}
\label{app:detailed_dynamics}

To understand the localized optimization mechanism, Figure~\ref{fig:detailed_dynamics} plots the evolution of the objective components over training for two distinct $n=6$ targets: a commutative group ($Z_6$) and a {non-group-isotope quasigroup} ($NG_2$).

\subsubsection{Group Isotope Dynamics ($Z_6$):} 
As shown in Figure~\ref{fig:detailed_dynamics} (Top), the parameters approach the collinear set as the alignment gap $\mathcal R_\delta \coloneqq \mathcal H- \mathcal B_\delta$ smoothly decays toward zero around epoch $t=400$.
After $\lambda$ is set to zero at epoch $t=500$, the total objective $\mathcal H$, the base penalty $\mathcal B_\delta$, and the output-dependent lower bound $\mathcal H^*_\delta$ approach the exact-fit floor $3|\supp(\delta)|=3\cdot6^2=108$, while $\mathcal R_\delta\approx0$.

\subsubsection{Non-Group Dynamics ($NG_2$):} 
\begin{figure*}[ht]
  \begin{center}
        \includegraphics[width=0.95\textwidth]{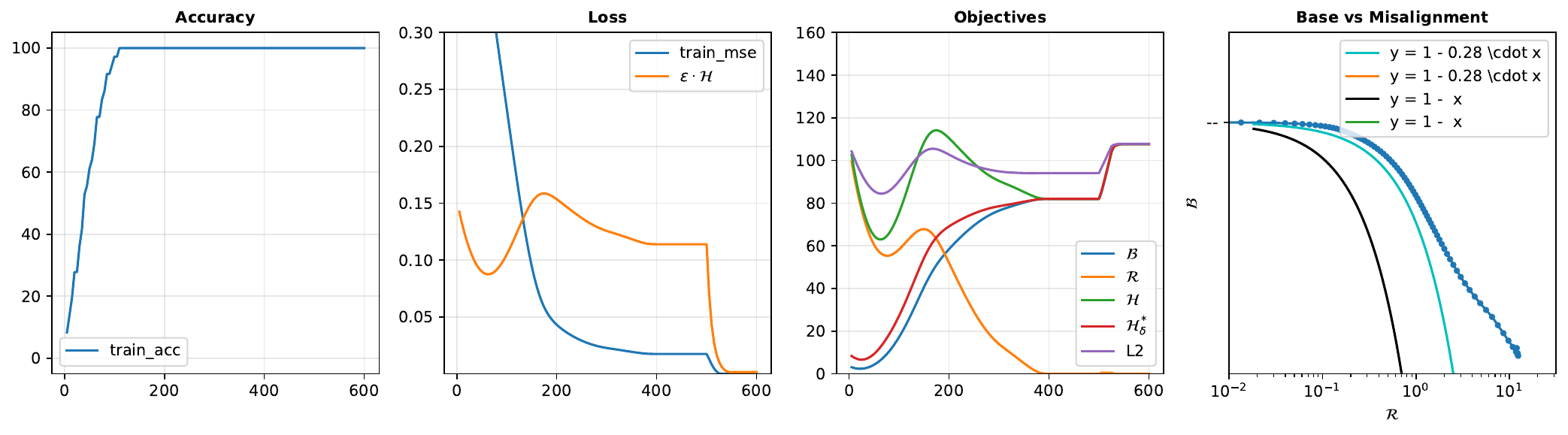}
    \vspace{0.2cm}
    
        \includegraphics[width=0.95\textwidth]{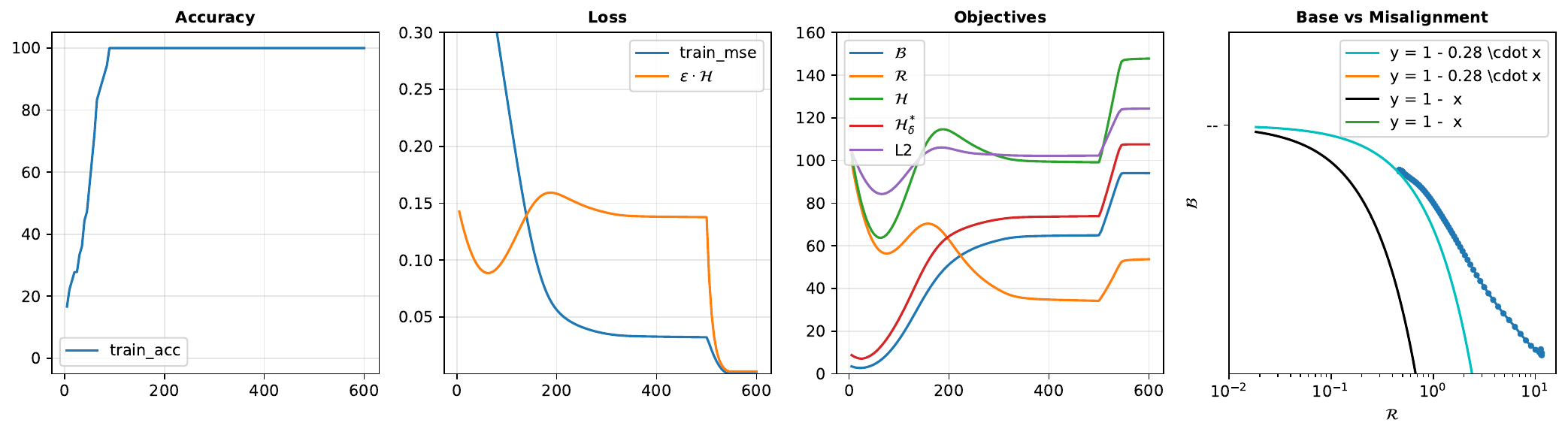}
  \end{center}
    \caption{
        \textbf{Component-wise Optimization Dynamics.}
        The evolution of the objective components ($\mathcal H$, $\mathcal B_\delta$, $\mathcal R_\delta$, $\mathcal H^*_\delta$) alongside the train loss and accuracy during optimization. The sensitivity coefficient $\lambda$ is held at $0.05$ until epoch $500$, and then dropped to $0$ to fit the target exactly.
        \textbf{(Top)} For the $Z_6$ group isotope, the misalignment $\mathcal R_\delta$ {decays to zero}. In the exact-fit phase ($t>500$), the components ($\mathcal H$, $\mathcal B_\delta$, $\mathcal H^*_\delta$) approach the $3|\supp(\delta)|=108$ theoretical floor.
        \textbf{(Bottom)} For the $NG_2$ non-group target, {the observed endpoint retains positive misalignment}. At the training endpoint, the total objective $\mathcal H$ is elevated above the bound, visualizing  observed {complexity excess}. 
    }
  \label{fig:detailed_dynamics}
\end{figure*}

Conversely, Figure~\ref{fig:detailed_dynamics} (Bottom) illustrates the structural obstruction faced by non-group-isotope targets. Because the feasible collinear set is empty for $NG_2$, {exact feasible reconstruction cannot have zero alignment gap; in the training run,} the alignment gap $\mathcal R_\delta$ plateaus at a positive value. After setting $\lambda=0$ at epoch $500$, the {bound} $\mathcal H^*_\delta$ {approaches} $108$. The total objective is $\mathcal H\approx145$ at the training endpoint.
The endpoint therefore has an observed {complexity excess} of approximately $37$ above the {exact-fit floor}, alongside an alignment gap $\mathcal R_\delta\approx55$.

\begin{table}[ht]
\centering
\caption{Cayley Table for $Z_2 \times Z_3 \cong Z_6$ (Group Isotope, zero latent associativity violation)}
\label{tab:cayley_z6}
\begin{tabular}{c|cccccc}
$\circ$ & 0 & 1 & 2 & 3 & 4 & 5 \\ \hline
0 & 0 & 1 & 2 & 3 & 4 & 5 \\
1 & 1 & 0 & 3 & 2 & 5 & 4 \\
2 & 2 & 3 & 4 & 5 & 0 & 1 \\
3 & 3 & 2 & 5 & 4 & 1 & 0 \\
4 & 4 & 5 & 0 & 1 & 2 & 3 \\
5 & 5 & 4 & 1 & 0 & 3 & 2
\end{tabular}
\end{table}

\begin{table}[ht]
\centering
\caption{Cayley Table for a Non-Group Isotope: Latent associativity violation count = 80}
\label{tab:cayley_ng2}
\begin{tabular}{c|cccccc}
$\circ$ & 0 & 1 & 2 & 3 & 4 & 5 \\ \hline
0 & 0 & 1 & 2 & 3 & 4 & 5 \\
1 & 1 & 0 & 3 & 4 & 5 & 2 \\
2 & 2 & 4 & 5 & 0 & 1 & 3 \\
3 & 3 & 5 & 4 & 2 & 0 & 1 \\
4 & 4 & 2 & 1 & 5 & 3 & 0 \\
5 & 5 & 3 & 0 & 1 & 2 & 4
\end{tabular}
\end{table}

\subsection{Population Results and the Global Landscape}
\label{app:population_results}

{ We extend the individual examples above to the population studied in Section~\ref{sec:empirical}, comparing normalized optimization trajectories and training endpoints with the target's latent associativity violation (Figure~\ref{fig:associativity_violation}).}

\paragraph{Experimental Setup.}
We analyze reduced Latin squares $\delta$ corresponding to loops of orders $n \in \{5, 6, 7, 8\}$. For the smaller orders $n \in \{5, 6\}$, we exhaustively evaluate all unique loops up to isomorphism (6 and 109 cases, respectively). For $n \in \{7, 8\}$, we evaluate a random subset of 100 loops per order, due to the combinatorial explosion of the search space (e.g., 23,746 unique loops for $n=7$). 

For each $\delta$, we use full-batch gradient descent with multiple random initializations and record the attained endpoint objective values, denoting a training endpoint by $\Theta_\infty$. Let $\mathfrak L(\delta)$ be the set of reduced loop tables isotopic to $\delta$. To correlate the landscape geometry with algebraic structure, we define the normalized latent associativity violation count
\[
\nu(\delta)\coloneqq
\min_{(Q,\star)\in\mathfrak L(\delta)}\frac{1}{|\supp(\delta)|}
\sum_{a,b,c\in Q}
\mathbb I_{\{(a\star b)\star c\ne a\star(b\star c)\}}.
\]
Every quasigroup is isotopic to a loop, so $\mathfrak L(\delta)$ is nonempty. If $\delta$ and $\delta'$ are isotopic, then $\mathfrak L(\delta)=\mathfrak L(\delta')$; consequently, $\nu(\delta)=\nu(\delta')$. Moreover, $\nu(\delta)=0$ if and only if $\delta$ is a group isotope.

{ 
\paragraph{Normalized Trajectories and Bounds.}
Following Section~\ref{sec:empirical}, we normalize the sensitivity and the Cauchy--Schwarz bound by the output-dependent lower bound:
\[
\widetilde{\mathcal H}_\delta\coloneqq\frac{\mathcal H}{\mathcal H^*_\delta(T(\Theta))},\qquad
\widetilde{\mathcal B}_\delta\coloneqq\frac{\mathcal B_\delta}{\mathcal H^*_\delta(T(\Theta))},\qquad
\widetilde{\mathcal R}_\delta\coloneqq\widetilde{\mathcal H}_\delta-\widetilde{\mathcal B}_\delta.
\]
The left panel of Figure~\ref{fig:associativity_violation} shows the full trajectories in the $(\widetilde{\mathcal R}_\delta,\widetilde{\mathcal B}_\delta)$ plane, and the middle panel enlarges the training endpoints. From random initialization, the trajectories sweep right-to-left from a high normalized alignment gap toward smaller $\widetilde{\mathcal R}_\delta$ and larger $\widetilde{\mathcal B}_\delta$.

The red contour marks $\widetilde{\mathcal H}_\delta=\widetilde{\mathcal R}_\delta+\widetilde{\mathcal B}_\delta=1$. Theorem~\ref{thm:unconditional_bound} gives $\widetilde{\mathcal H}_\delta\geq1$ globally, including outside the collinear set, and the plotted states respect this bound. The cyan dotted reference $\widetilde{\mathcal B}_\delta=1-0.28\widetilde{\mathcal R}_\delta$ corresponds to $\widetilde{\mathcal H}_\delta-1=0.72\widetilde{\mathcal R}_\delta$ along that line. It suggests a stronger, alignment-dependent lower bound: the observed normalized excess complexity grows with the normalized alignment gap.

\paragraph{Training Endpoints and the Associativity Gap.}
The training endpoints are consistent with the floor-attainment characterization and positive infimum gap of Theorem~\ref{thm:global_optimality_quasigroup}. For group-isotope targets ($\nu(\delta)=0$), they approach the floor point $(\widetilde{\mathcal R}_\delta,\widetilde{\mathcal B}_\delta)=(0,1)$, where $\widetilde{\mathcal H}_\delta=1$. For targets not isotopic to a group, the endpoints retain a positive alignment gap and complexity above the universal floor. The theory excludes feasible collinear factorizations for these targets and establishes a positive infimum gap.

Across the tested targets, larger $\nu(\delta)$ is associated with both larger normalized alignment gaps and higher attained normalized complexity $\widetilde{\mathcal H}_\delta(\Theta_\infty)$, connecting the middle and right panels of Figure~\ref{fig:associativity_violation}. This empirical association suggests a graded relationship beyond the theorem's zero-versus-positive gap characterization. Both $\mathcal H_{\inf}$ and $\nu$ are isotopy-invariant; the plotted complexity values are attained training outcomes.
}

{
\section{ Arbitrary Binary Operations: Beyond Quasigroups}
\label{app:beyond-quasigroups}

{Let $Q$ be a nonempty finite set of cardinality $n$, and let $\circ:Q\times Q\to Q$ be any binary operation. Set $\delta_{abc}=\mathbb I_{\{a\circ b=c\}}$.} We use the normalized matrix norm and inner product of the main text, full feasibility $T_{abc}=\delta_{abc}$ for all triples, and the support-weighted objective \eqref{eq:H-Latin}. {No cancellation, associativity, or surjectivity of $\circ$ is assumed.}

Write $U_\circ\coloneqq\{a\circ b:a,b\in Q\}$ for the used-output set.
By Theorem~\ref{thm:absolute_lower_bound} and its proof, a feasible factorization attains $3 \, |\supp(\delta)|$ if and only if all input slices and all used output slices are unitary and
\begin{equation}
\label{eq:general-supported-product}
A_aB_bC_{a\circ b}=I_n,\qquad A_aB_b=C_{a\circ b}^\dagger
\quad (a,b\in Q).
\end{equation}
No unitarity condition is imposed on unused output slices.

\subsection{Collision Propagation and Positive Complexity Gaps}

Define equivalence relations for the two input roles by
\begin{align*}
a\sim_R a'&\quad\Longleftrightarrow\quad \forall b\in Q,\ {a\circ b=a'\circ b},\\
b\sim_C b'&\quad\Longleftrightarrow\quad \forall a\in Q,\ {a\circ b=a\circ b'}.
\end{align*}
Repeated rows (columns) mean that $\sim_R$ ($\sim_C$) relates distinct labels.

\begin{theorem}[Input redundancy and consistent collisions]
\label{thm:general-collision-rigidity}
Suppose $\Theta\in\mathcal F_\delta$ and $\mathcal H(\Theta)=3 \, |\supp(\delta)|$. Then
\[
A_a=A_{a'}\ \Longleftrightarrow\ a\sim_R a',\qquad
B_b=B_{b'}\ \Longleftrightarrow\ b\sim_C b'.
\]
Moreover, for every fixed $a,b\in Q$,
\begin{align*}
{a\circ b=a\circ b'}&\quad\Longleftrightarrow\quad b\sim_C b',\\
{a\circ b=a'\circ b}&\quad\Longleftrightarrow\quad a\sim_R a'.
\end{align*}
\end{theorem}
\begin{proof}
If $a\sim_R a'$, choose any $b$. Equation~\eqref{eq:general-supported-product} gives $A_aB_b=A_{a'}B_b$; cancel the unitary $B_b$ to obtain $A_a=A_{a'}$. Conversely, equality of these matrices gives equality of $T_{abc}$ and $T_{a'bc}$ for every $b,c$. Full one-hot feasibility then gives identical rows. The proof for columns is symmetric.

If {$a\circ b=a\circ b'$}, Equation~\eqref{eq:general-supported-product} gives $A_aB_b=A_aB_{b'}$. Canceling $A_a$ yields $B_b=B_{b'}$, hence identical columns by the first assertion. Conversely, identical columns give the displayed collision. The other input role follows by canceling $B_b$.
\end{proof}

\begin{theorem}[Group isotope or input redundancy]
\label{thm:general-floor-structure}
If a feasible factorization attains $3 \, |\supp(\delta)|$, then {$(Q,\circ)$} is a group isotope, or it has repeated rows, or it has repeated columns. In particular, if all rows and columns are distinct, floor attainment implies that {$(Q,\circ)$} is a quasigroup and a group isotope. The same implication holds if both input-slice maps $a\mapsto A_a$ and $b\mapsto B_b$ are injective.
\end{theorem}
\begin{proof}
With no repeated columns, the first collision equivalence in Theorem~\ref{thm:general-collision-rigidity} makes every map {$b\mapsto a\circ b$} injective, hence bijective on the finite set $Q$. With no repeated rows, every map {$a\mapsto a\circ b$} is likewise bijective. Thus {$(Q,\circ)$} is a quasigroup. Its floor-attaining factorization is unitary and collinear by Theorem~\ref{thm:absolute_lower_bound}; the quasigroup characterization (Theorem~\ref{thm:unitary_equivalence}) implies group isotopy. 
If either distinctness assumption fails, the target has repeated rows or columns, proving the first assertion.
Injectivity of the two input-slice maps excludes repeated rows and columns by Theorem~\ref{thm:general-collision-rigidity}.
\end{proof}

{
\begin{corollary}[Uniform infimum gaps from stray collisions or no redundancy]
\label{cor:general-stray-collision}
If some $a,a',b,b'$ satisfy
\[
{a\circ b=a\circ b',\qquad a'\circ b\ne a'\circ b'},
\]
then there exists {$\varepsilon_\circ > 0$} such that every fully feasible factorization satisfies {$\mathcal H(\Theta) \geq 3 \, |\supp(\delta)| + \varepsilon_\circ$}. The same uniform-gap assertion holds for a collision in one column that fails to persist in another column, and for a non-group-isotope operation with all rows and columns distinct.
\end{corollary}
\begin{proof}
Equality at the floor would propagate the indicated collision to every row by Theorem~\ref{thm:general-collision-rigidity}, contradicting the disagreement. The symmetric case follows from column collision propagation. For a non-group-isotope operation with distinct rows and columns, floor attainment contradicts Theorem~\ref{thm:general-floor-structure}. Each case excludes every feasible floor point, so Theorem~\ref{thm:positive-infimum-gap} supplies the positive uniform gap.
\end{proof}
}

\subsection{The Role-Separated Reduced Table}

Let {$R_\circ=Q/{\sim_R}$, $S_\circ=Q/{\sim_C}$}, and retain the used-output set {$U_\circ$}. Define
\[
{\bar{\circ}:R_\circ\times S_\circ\longrightarrow U_\circ,\qquad
[a]_R\mathbin{\bar{\circ}}[b]_C=a\circ b.}
\]
This construction keeps the three roles separate.

\begin{theorem}[Reduced-table cancellation and cardinalities]
\label{thm:general-reduced-table}
For every operation, {$\bar{\circ}$} is well-defined and surjective. If a feasible factorization attains the floor, {$\bar{\circ}$} is injective in each argument separately. Writing {$r=|R_\circ|$, $s=|S_\circ|$, and $t=|U_\circ|$}, one has
\[
\max(r,s) \leq  t \leq  rs.
\]
\end{theorem}
\begin{proof}
Replacing $a$ by an equivalent row representative leaves every entry unchanged; replacing $b$ by an equivalent column representative does the same. Thus {$\bar{\circ}$} is well-defined. Every used output has an input pair, proving surjectivity and $t \leq  rs$.

At the floor, equal outputs in a fixed row imply identical columns by Theorem~\ref{thm:general-collision-rigidity}, hence equal column classes. This proves injectivity in the second argument, and the symmetric reasoning proves injectivity in the first. Since both class sets are nonempty, fixing a column class embeds {$R_\circ$ into $U_\circ$}, and fixing a row class embeds {$S_\circ$ into $U_\circ$}. Thus $r \leq  t$ and $s \leq  t$.
\end{proof}

\begin{theorem}[Used-output orthonormality]
\label{thm:general-output-orthonormality}
At a feasible floor-attaining factorization, the used output slices satisfy
\[
\langle C_c,C_d\rangle=\Tr(C_c^\dagger C_d)
=\mathbb I_{\{c=d\}},\qquad c,d\in {U_\circ}.
\]
\end{theorem}
\begin{proof}
Choose $a,b$ with {$a\circ b=c$}. Equation~\eqref{eq:general-supported-product} and full feasibility give
$\Tr(C_c^\dagger C_d)=\Tr(A_aB_bC_d)=\delta_{abd}=\mathbb I_{\{c=d\}}$.
\end{proof}

\begin{proof}[Proof of Theorem~\ref{thm:exact-reduced-factorization}]
At a feasible floor point, Theorem~\ref{thm:general-collision-rigidity} makes $A$ and $B$ constant on their respective equality classes. They therefore define $U$ and $V$, while $W_c=C_c$ for $c\in U_\circ$. Theorem~\ref{thm:absolute_lower_bound} and \eqref{eq:general-supported-product} give unitarity and the product identity, and Theorem~\ref{thm:general-output-orthonormality} gives trace orthogonality.
Conversely, set $A_a=U_{[a]_R}$, $B_b=V_{[b]_C}$, and $C_c=W_c$ for used outputs, with $C_c=0$ otherwise. For used $c$, the product identity and orthogonality give
$T_{abc}=\Tr(W_{a\circ b}^\dagger W_c)=\mathbb I_{\{a\circ b=c\}}$; unused outputs give zero. Thus the factorization is fully feasible, and its supported unitary products are identity, so Theorem~\ref{thm:absolute_lower_bound} yields $\mathcal H=3 \, |\supp(\delta)|$.

Output orthogonality makes $W$ injective. If $U_r=U_{r'}$, then the product identity and injectivity of $W$ give $r\mathbin{\bar\circ}s=r'\mathbin{\bar\circ}s$ for every $s$. Representatives of $r,r'$ therefore have identical entire rows, so $r=r'$. The column argument is symmetric, proving injectivity of $V$. Consequently, the distinct input matrices are in bijection with the row and column classes.
\end{proof}

\paragraph{Structural and coding interpretation.}
The theorem separates two levels of representation: the input maps merge precisely globally indistinguishable symbols, while the reduced factors distinguish every row class, column class, and used output. Multiplication by a fixed unitary then preserves these distinctions, ruling out context-dependent merging in either input role. This is a constraint on symbolic composition, even though every participating matrix has full rank.

The encoder cardinality bounds following Theorem~\ref{thm:exact-reduced-factorization} use only exact decoding: if $e_R(a)=e_R(a')$, exact decoding implies $a\circ b=D(e_R(a),e_C(b))=D(e_R(a'),e_C(b))=a'\circ b$ for every $b$. Thus equal first-input encoder outputs imply identical rows, and choosing one representative per row class gives an injection into $E_R$. The same argument gives an injection from $S_\circ$ into $E_C$. Hence $|R_\circ|\leq|E_R|$ and $|S_\circ|\leq|E_C|$, independently of floor attainment.
The canonical quotient encoders attain both bounds, with the reduced table as decoder. Consequently, fixed-length binary encodings require at least $\lceil\log_2|R_\circ|\rceil$ and $\lceil\log_2|S_\circ|\rceil$ bits, respectively, and binary labels for the quotient classes achieve these lengths. 
Their optimality concerns separate input-state alphabets, with an unrestricted decoder; the matrix representations supply exact operator composition on those alphabets.

\subsection{Finite-Group Realizations}
\label{app:finite-group-realizations}

The unitary realization forced by floor attainment has a finite-group counterpart. As in Theorem~\ref{thm:exact-reduced-factorization}, let $R_\circ,S_\circ,U_\circ$ be the reduced row, column, and output sets.

\begin{corollary}[Finite-Group Realizations at the Floor]
\label{cor:floor-finite-group-code}
Every floor-attaining operation admits a finite group $G$ and injections $\alpha:R_\circ\to G$, $\beta:S_\circ\to G$, and $\gamma:U_\circ\to G$ satisfying
\[
\alpha(i)\beta(j)=\gamma(i\mathbin{\bar\circ}j).
\]
Thus its reduced table occurs as a subtable of a finite-group multiplication table after independent row, column, and output relabeling.
\end{corollary}
\begin{proof}
Theorem~\ref{thm:exact-reduced-factorization} gives injective unitary maps $U,V,W$ with $U_iV_j=W_{i\mathbin{\bar\circ}j}^\dagger$. The subgroup $\Gamma\leq\mathrm U(n)$ generated by these finitely many matrices is a finitely generated linear group. By Malcev's residual-finiteness theorem \citep{magnus1969residually}, each nonidentity element of $\Gamma$ has a nonidentity image in some finite quotient. Apply this to $X^{-1}Y$ for every pair of distinct matrices in the finite set of values of $U,V,W^\dagger$. Taking the product of the finitely many quotient maps gives a homomorphism $\pi:\Gamma\to G$ into a finite group that is injective on this set. Then $\alpha(i)=\pi(U_i)$, $\beta(j)=\pi(V_j)$, and $\gamma(k)=\pi(W_k^\dagger)$ are injective and satisfy the required product identity. If the set has only one element, the trivial quotient suffices.
\end{proof}

\paragraph{A non-quasigroup floor-attaining family.}
Let $G$ have order $m$, let $n=km$ for an integer $k\geq2$, and choose surjections $p,q:[n]\to G$ and an injection $\omega:G\to[n]$. The operation $a\circ b=\omega(p(a)q(b))$ has $r=s=t=m<n$. Taking $k$ copies of the unitary left-regular representation $\rho_G$ gives an $n$-dimensional representation $\rho$ with normalized trace $\Tr(\rho(g))=\mathbb I_{\{g=e\}}$. Set $A_a=\rho(p(a))$, $B_b=\rho(q(b))$, and $C_{\omega(g)}=\rho(g)^\dagger$, with unused output slices zero. These factors reconstruct the full target and attain the floor.

\paragraph{Group embeddability without floor attainment.}
On $Q=\{0,1,2\}$, define $a\circ b=(a+b)\bmod 2$. Rows $0$ and $2$ are identical, as are columns $0$ and $2$. Identifying each input class by its parity yields the two-element group table, with used outputs $\{0,1\}$. Thus the reduced table is group-embeddable and separately cancellative.

Suppose a feasible factorization attained $3 \, |\supp(\delta)|=27$. Theorem~\ref{thm:exact-reduced-factorization} would give unitary maps into $\mathrm U(3)$ satisfying
\[
U_0V_0=U_1V_1=W_0^\dagger,
\qquad
U_0V_1=U_1V_0=W_1^\dagger.
\]
Set $P=U_0^\dagger U_1=V_1V_0^\dagger$. The product identities and orthonormality of the output matrices imply
\[
P^2=U_0^\dagger U_1V_1V_0^\dagger=I_3,
\qquad
\Tr(P)=\Tr(W_1^\dagger W_0)=0.
\]
But a unitary involution has eigenvalues $\pm1$, whose sum cannot vanish in dimension three. This contradiction excludes floor attainment despite the group reduced table, demonstrating that group embeddability alone does not guarantee a unitary realization in the prescribed dimension with orthonormal output matrices.

\subsection{Examples Separating Associativity and Redundancy}
\label{app:extension-examples}

\paragraph{A floor-attaining operation with no associative isotope.}
On $Q=\{0,1,2,3\}$, let $p=(0,1,1,1)$ and $q=(0,0,1,1)$, and define {$a\circ b=2p(a)+q(b)$}. Its table is
\[
\begin{pmatrix}
0&0&1&1\\
2&2&3&3\\
2&2&3&3\\
2&2&3&3
\end{pmatrix}.
\]
Put $D=\operatorname{diag}(1,i,-1,-i)$ and choose $A_a=D^{2p(a)}$, $B_b=D^{q(b)}$, $C_c=D^{-c}$. Since $\Tr(D^k)$ is $1$ for $k=0 \pmod4$ and $0$ otherwise, these matrices give full one-hot feasibility. All slices are unitary and the supported products are $I_4$, so $\mathcal H=48$. The reduced table has $(r,s,t)=(2,2,4)$ despite full-rank latent matrices.

To see that no isotope is associative, write any isotope as $g(x,y)=h(P(x),S(y))$, where $P,S:Q\to\{0,1\}$ are surjective, $h:\{0,1\}^2\to Q$ is bijective, and the two fibers of $P$ have sizes $1$ and $3$. Associativity would give
\[
h(P(h(P(x),S(y))),S(z))
=h(P(x),S(h(P(y),S(z)))).
\]
Injectivity of $h$ in its pair of arguments forces $P(h(P(x),S(y)))=P(x)$ for all $x,y$. Surjectivity of $P,S$ then implies $P(h(j,k))=j$ for every $j,k\in\{0,1\}$. For each $j$, the two distinct labels $h(j,0),h(j,1)$ belong to its $P$-fiber. Both fibers must therefore have size $2$, contradicting their sizes $1$ and $3$. Hence even isotopy cannot make this floor-attaining operation associative.

\paragraph{An associative semigroup that cannot attain the floor.}
On $Q=\{0,1\}$, let {$a\circ b=\min(a,b)$}, with table
\[
\begin{pmatrix}0&0\\0&1\end{pmatrix}.
\]
This operation is associative, but { $0\circ0=0\circ1$ while $1\circ0\ne1\circ1$}. Corollary~\ref{cor:general-stray-collision} gives a constant { $\varepsilon_\circ>0$} with { $\mathcal H(\Theta) \geq 12 + \varepsilon_\circ > 12$} for every fully feasible factorization. Thus associativity alone is not sufficient for floor attainment outside the quasigroup setting. Its rows and columns are distinct: the stray collision is not identical-row or identical-column redundancy.

\paragraph{The floor does not determine redundancy counts.}
At the same order $n$, group operations and the left-zero operation {$a\circ b=a$} both attain $3 \, |\supp(\delta)|$, although their class counts are $(r,s)=(n,n)$ and $(n,1)$, respectively. For the latter, take the regular representation $\rho$ of a cyclic group of order $n$ and set $A_a=\rho(a)$, $B_b=I_n$, $C_c=\rho(c)^\dagger$. The regular character gives full feasibility and all supported products are identity. This illustrates why combinatorial redundancy and latent matrix rank should be treated as distinct complexity coordinates rather than inferred from a single objective value.

}

\section{A Description-Length Consequence}
\label{app:description-length}

For labeled tables on $[n]$, $n\geq2$, use a fixed computable lossless prefix decoder $D$ supporting group-isotope and literal-table formats, and write $\ell_D(\circ)$ for its shortest description length in bits. All logarithms are base two. A group isotope is determined by a group isomorphism type and three permutations of $[n]$. A computable enumeration of group types and permutation indices therefore gives
\begin{equation}
\label{eq:group-isotope-code-length}
\ell_D(\circ)\leq\lceil\log g(n)\rceil+3\lceil\log(n!)\rceil+O(\log n)
=O(n\log n),
\end{equation}
where $g(n)$ counts group isomorphism types and $\log g(n)=O((\log n)^3)$ by \citet{pyber1993enumerating}. The order header is self-delimiting; decoding time is unrestricted. In contrast, the Latin-square count satisfies $\log L_n=n^2\log n+O(n^2)$ \citep[Section~7]{mckay2005latin}. Counting binary descriptions shows that almost all quasigroup tables require $\Omega(n^2\log n)$ bits under any fixed decoder; a literal-table fallback gives $\Theta(n^2\log n)$ for typical tables under the same decoder.

Thus, the quasigroup class attaining the sensitivity floor admits economical descriptions. The replicated family above also admits $O(n\log n)$-bit descriptions by encoding its group, class maps, and output labels.

\end{document}